\documentclass[10pt,onecolumn,letterpaper]{article}

\usepackage[pagenumbers]{cvpr} 
\usepackage[utf8]{inputenc}

\usepackage{graphicx}
\usepackage{amsmath}
\usepackage{amssymb}
\usepackage[accsupp]{axessibility}  
\usepackage{bm}
\usepackage{booktabs}
\usepackage{multicol}
\usepackage{multirow}
\usepackage{amssymb}
\usepackage{soul}
\usepackage{appendix}
\usepackage{arydshln}
\usepackage{colortbl}
\usepackage{framed}
\usepackage{fancyvrb} 
\usepackage{listings} 
\usepackage{xcolor} 
\usepackage{tabularx} 
\usepackage{amsfonts}
\usepackage{algorithm}
\usepackage{algpseudocode}
\usepackage{amsmath}
\newcolumntype{L}[1]{>{\raggedright\arraybackslash}p{#1\textwidth}}
\newcolumntype{C}[1]{>{\centering\arraybackslash}p{#1\textwidth}}
\newcolumntype{R}[1]{>{\raggedleft\arraybackslash}p{#1\textwidth}}

\definecolor{keywordcolor}{RGB}{0, 0, 255} 
\definecolor{stringcolor}{RGB}{255, 140, 0} 
\definecolor{commentcolor}{RGB}{0, 128, 0} 

\newcommand{\minisection}[1]{\vspace{1mm}\noindent{\textbf{#1}~}}

\usepackage[pagebackref,breaklinks,colorlinks]{hyperref}

\usepackage[capitalize]{cleveref}
\usepackage{cite}
\usepackage{subcaption}

\crefname{section}{Sec.}{Secs.}
\Crefname{section}{Section}{Sections}
\Crefname{table}{Table}{Tables}
\crefname{table}{Tab.}{Tabs.}

\begin{document}

\title{LLM-AutoDiff: Auto-Differentiate Any LLM Workflow}  


\author{
    Li Yin$^1$, Zhangyang ``Atlas" Wang$^2$\vspace{0.3em}\\
    $^1$SylphAI \quad $^2$University of Texas at Austin\vspace{0.5em}\\
    }
\maketitle
\graphicspath{ {./fig/} }

\begin{abstract}
Large Language Models (LLMs) have reshaped natural language processing, powering applications from multi-hop retrieval and question answering to autonomous agent workflows. Yet, \emph{prompt engineering}—the task of crafting textual inputs to effectively direct LLMs—remains difficult and labor-intensive, particularly for complex pipelines that combine multiple LLM calls with functional operations like retrieval and data formatting. We introduce \textbf{LLM-AutoDiff}: a novel framework for Automatic Prompt Engineering (APE) that extends textual gradient-based methods (such as \emph{Text-Grad}) to multi-component, potentially cyclic LLM architectures. Implemented within the \textit{AdalFlow} library,\footnote{\url{https://github.com/SylphAI-Inc/AdalFlow}} LLM-AutoDiff treats each textual input as a trainable parameter and uses a frozen ``backward engine'' LLM to generate feedback—akin to ``textual gradients''—that guide iterative prompt updates. Unlike prior single-node approaches, LLM-AutoDiff inherently accommodates functional nodes, preserves time-sequential behavior in repeated calls (e.g., multi-hop loops), and combats the ``lost-in-the-middle'' problem by isolating distinct sub-prompts (instructions, formats, or few-shot examples). It further boosts training efficiency by focusing on error-prone samples through selective gradient computation. Across diverse tasks, including single-step classification, multi-hop retrieval-based QA, and agent-driven pipelines, LLM-AutoDiff consistently outperforms existing textual gradient baselines in both accuracy and training cost. By unifying prompt optimization through a graph-centric lens, LLM-AutoDiff offers a powerful new paradigm for scaling and automating LLM workflows — mirroring the transformative role that automatic differentiation libraries have long played in neural network research.

\begin{figure}[t]
      \centering
\includegraphics[width=1\textwidth]{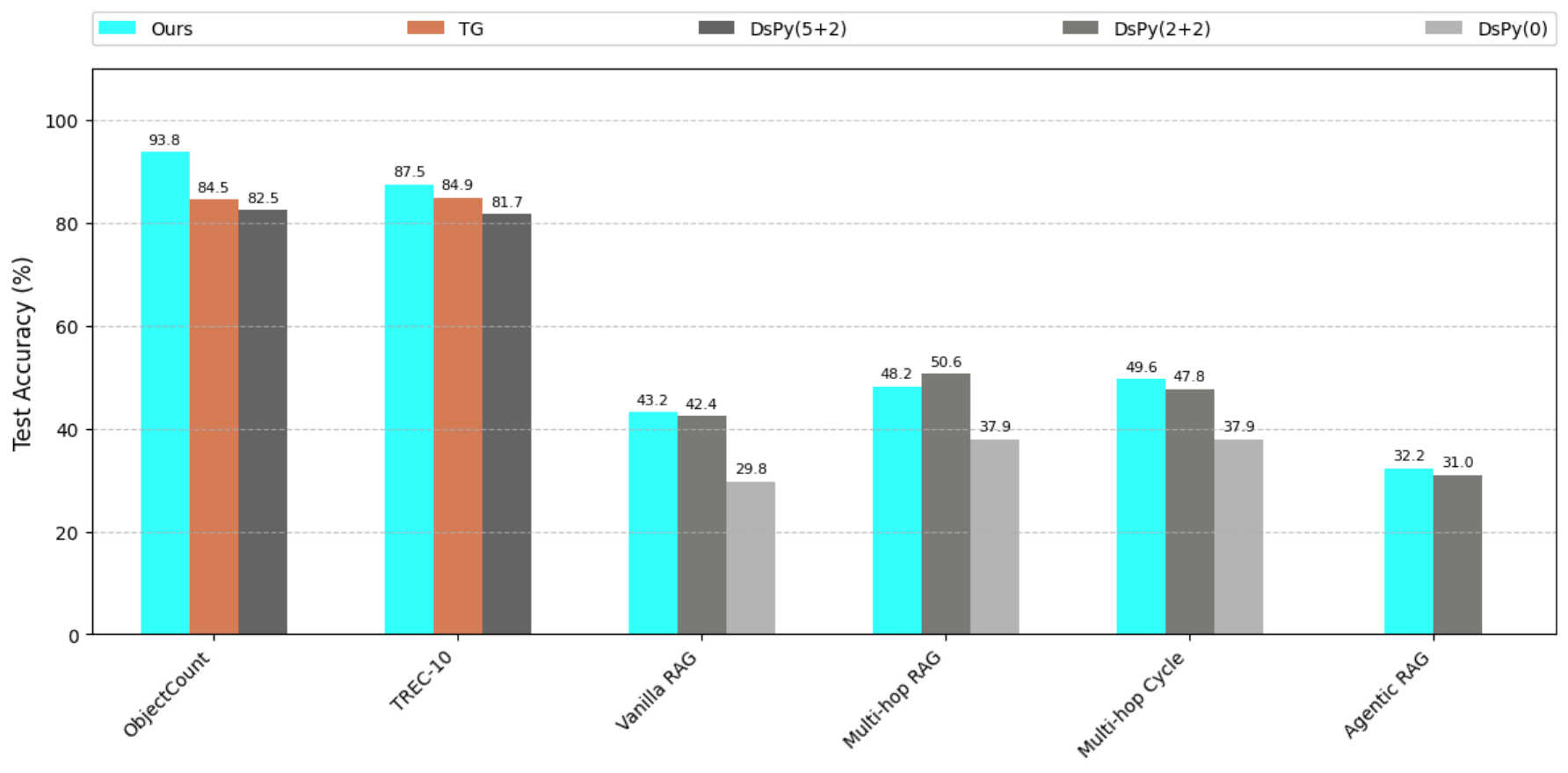}
      \caption{
\textbf{Benchmark Performance of LLM-AutoDiff in AdalFlow.}
}

      \label{fig:auto_graph_agent}

\end{figure}
\end{abstract}

\section{Introduction}

\label{sec:intro}

Large Language Models (LLMs)~\cite{brown2020language} lay the groundwork for a wide array of applications: from consumer-facing chatbots to academic research prototypes, LLM-driven systems are increasingly defining both front-end interfaces and back-end logic. This rapid progress has highlighted an important focal point in LLM-based development: the design of \emph{prompts}, which are sequences of text that specify tasks, include examples, or provide instructions for generating the desired output. Often referred to as \emph{prompt engineering}~\cite{wei2022chain, shinn2024reflexion}, this skillful curation of textual inputs can dramatically impact performance, as subtle variations in phrasing or structure can change model behavior in unpredictable ways.

Although prompt engineering is effective for single-step tasks, its limitations become pronounced when dealing with complex architectures composed of multiple LLM modules and auxiliary operations. Systems that rely on multi-hop retrieval~\cite{khattab2022demonstrate} or autonomous agent loops~\cite{yao2022react,kimllm} must orchestrate not only the LLM calls themselves, but also the data flow among external processes such as retrievers, validators, and deduplicators. The result is a network of interconnected components that can contain cycles, conditional branches, and iterative procedures. Manual prompt tuning for each module in such a network can be both time-consuming and error-prone. As a result, developers often encounter the frustration of having to guess the right textual formulations or re-verify prompts whenever a sub-component, dataset, or even the underlying LLM changes.

\begin{figure}[t]
      \centering
\includegraphics[width=1\textwidth]{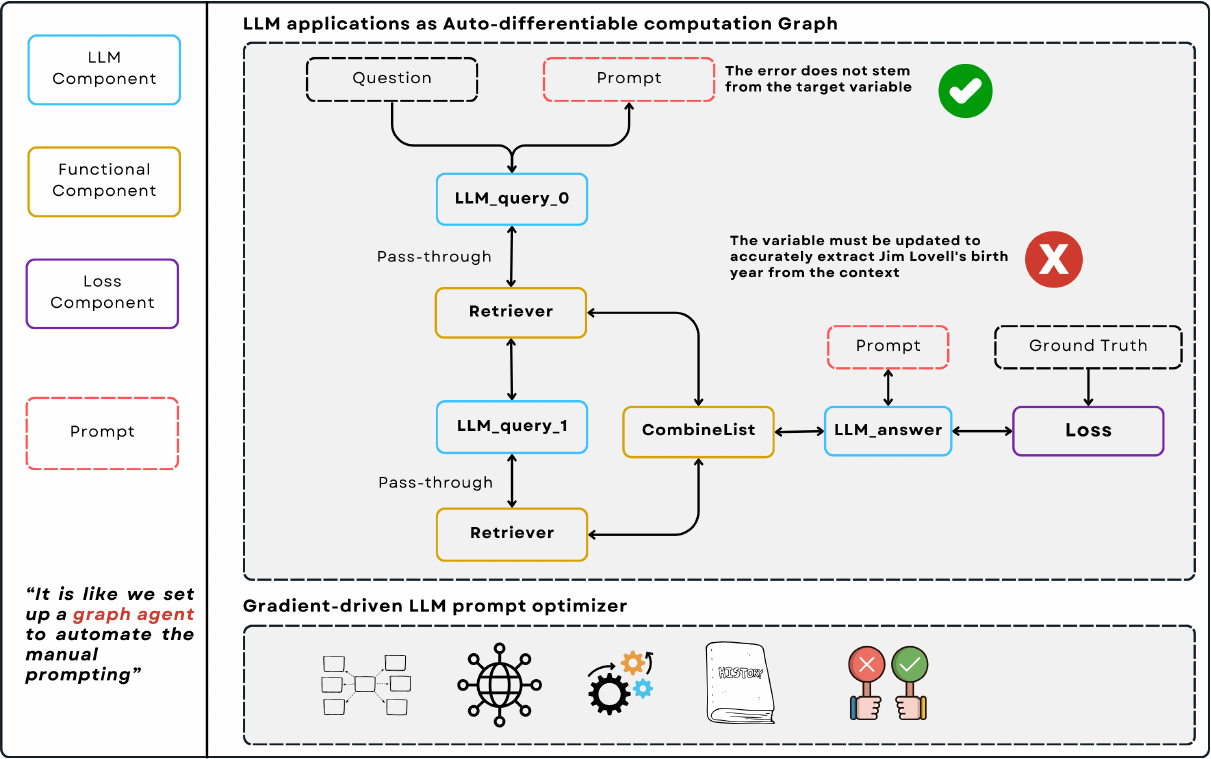}
      \caption{
\textbf{Representing an LLM application as an auto-differentiable computation graph}. We illustrate how each node in the graph can be one of three types: an \textit{LLM component}, a \textit{functional component}, or a \textit{loss component}. 
By coupling this graph with a gradient-driven LLM prompt optimizer, the traditionally labor-intensive task of manually crafting prompts is automated. 
During training, a forward pass traces every intermediate input and output, while the backward pass backpropagates ``evaluation'' signals (e.g., losses or textual gradients) through the network to locate and correct errors. 
Crucially, multiple invocations of the same node (such as when \textit{LLM\_query\_0} and \textit{LLM\_query\_1} are the same component called twice, along with the retriever being called twice) can accumulate multiple gradients, making it essential for the auto-differentiation engine to preserve the correct sequence of updates. 
This unified approach not only highlights where mistakes arise in multi-component pipelines (including loops), but also provides a systematic means to refine all relevant prompt parameters in tandem.
}

      \label{fig:auto_graph_agent}

\end{figure}

A growing body of research into \emph{Automatic Prompt Engineering (APE)} has sought to alleviate this burden by systematically refining prompts through automated methods~\cite{yang2023large, pryzant2023automatic, yuksekgonul2024textgrad}. A key innovation in this area is the concept of \emph{textual gradients}, which mimic numeric gradient descent by using a “backward engine” LLM to critique the output and propose edits that reduce a textual “loss.” Current approaches such as \emph{Text-Grad}~\cite{pryzant2023automatic, yuksekgonul2024textgrad} have validated the viability of this technique, demonstrating performance gains in tasks where a single LLM node is trained on labeled data via an iterative prompt-update loop. However, \textbf{these methods do not fully address the intricacies of multi-component or cyclic pipelines}, where prompts for different nodes may require simultaneous adaptation. Moreover, the interplay of functional operations—ranging from chunking retrieved documents to formatting data—introduces dependencies that these frameworks typically do not address.

\paragraph{Contributions.}
We present \textbf{LLM-AutoDiff} to close the gap in automatic prompt optimization for complex LLM pipelines, particularly those involving multi-component or cyclic structures. Our framework builds on the notion of \emph{textual gradients} yet \emph{significantly extends} single-node approaches by handling entire pipelines where each node—whether an LLM or a functional operation—can dynamically receive, modify, or relay textual instructions in response to downstream errors. Concretely, LLM-AutoDiff represents an LLM application as a directed graph that captures the flow of data and dependencies among LLMs, retrieval or formatting modules, and specialized evaluation routines. Within this graph, every textual input to an LLM is treated as a trainable parameter, allowing feedback signals to propagate not only into standard LLM prompts but also through nodes that themselves have no text to optimize (e.g., retrieval or deduplication). This “pass-through” mechanism ensures that any part of the pipeline contributing to errors can be identified and adjusted for improved final outputs.

Alongside these pass-through gradients, LLM-AutoDiff also introduces \emph{time-sequential gradients}, which are crucial for multi-hop reasoning or agent loops that repeatedly invoke certain LLM nodes. By attaching a time-stamped record to each invocation, the system delivers feedback in the correct temporal order, preventing confusion when the same node is called multiple times in a single run. Another key innovation is the treatment of different sub-prompts (such as task instructions, output formats, and few-shot examples) as distinct \emph{peers}, thereby mitigating the “lost-in-the-middle” effect~\cite{liu2024lost} and allowing the backward engine to pinpoint precisely which sub-prompt might have caused an error. Moreover, LLM-AutoDiff applies \emph{selective gradient computation} to concentrate updates on samples where the model’s outputs are incorrect, thus reducing token usage and training overhead.

To demonstrate the effectiveness of this framework, we run experiments on a broad spectrum of benchmarks. We start with simpler tasks (e.g., single-node classification and question answering) to show that LLM-AutoDiff either equals or surpasses contemporary textual gradient baselines in both accuracy and token efficiency. We then escalate to more advanced, multi-step scenarios where different prompts govern sequential retrieval-and-generation steps, including multi-node RAG setups, multi-hop RAG~\cite{lewis2020retrieval, khattab2022demonstrate}, and trainable ReAct agents~\cite{yao2022react} (with or without trainable LLM modules). In all these settings, LLM-AutoDiff avoids the redundancy of naive prompt search by unifying error signals and streamlining feedback delivery. Our findings also highlight the importance of respecting time-sequential feedback: ignoring the exact order of sub-queries can lead to ambiguity and suboptimal updates, whereas our time-stamped gradients keep improvements on track. Furthermore, by leveraging the few-shot optimization capability in \emph{AdalFlow}, we exceed existing approaches on final accuracy, inference speed, and training cost, all while preserving scalability and ease of use.

\textbf{In summary, LLM-AutoDiff} operationalizes the vision of \emph{Automatic LLM Application Optimization (ALAO)}, in which not only a single prompt, but also the entire network of prompts and operations in a pipeline can be refined automatically. Drawing direct inspiration from auto-differentiation frameworks like PyTorch~\cite{paszke2019pytorch} and TensorFlow~\cite{abadi2016tensorflow}, our system:
\begin{itemize}
    \item \textbf{Models an LLM-based pipeline as a directed (possibly cyclic) graph}, where each node may be a trainable LLM or a functional (non-trainable) operation.
    \item \textbf{Builds a runtime “parameter graph”} that tracks textual inputs (prompts, few-shot examples, etc.) as \emph{trainable parameters} amenable to iterative optimization.
    \item \textbf{Facilitates a textual backward pass} by using a frozen backward engine LLM to propagate feedback on errors—such as low accuracy—thus enabling an optimizer LLM to revise sub-prompts in a directed, gradient-like manner.
\end{itemize}
We further highlight five technical advances enabling and driving our system design:
\begin{enumerate}
    \item \textbf{Pass-Through Gradients for Functional Nodes:} Ensures gradients can flow through intermediary operations (e.g., context retrieval, deduplication) that manage data but have no learnable parameter nor prompt to directly update.
    \item \textbf{Time-Sequential Gradients for Cyclic Structures:} Accumulates multiple gradients for iterative or multi-hop calls, retaining the chronological order to avoid conflated feedback.
    \item \textbf{Peer Nodes for Sub-Prompt Clarity:} Splits instructions, format specifications, and few-shot examples into separate parameters to localize feedback and reduce confounding interactions.
    \item \textbf{Selective Gradient Computation (Error-Only):} Allocates compute and token usage where it is most needed—on incorrect samples—thereby improving efficiency and convergence speed.
    \item \textbf{Gradient-Driven Prompt Optimizer:} Extends OPRO~\cite{yang2023large} with prompt engineering heuristics and peer awareness, enabling multi-node systems to receive coherent and targeted updates from the optimizer.
\end{enumerate}
Taken together, these innovations unify prompt-level refinements under a single, end-to-end optimization process, offering a powerful new paradigm for large-scale LLM development.

\section{Related Work}
\label{sec:related}
This section first provides a broad view of traditional prompt engineering paradigms, then discusses the limitations of manual prompt design (Section~\ref{subsec:prompt_engineering}). We next review the growing landscape of \emph{Automatic Prompt Engineering (APE)} (Section~\ref{subsec:ape}), including single-node and advanced approaches. Finally, we highlight why a unified, multi-node perspective is needed to handle the full complexity of modern LLM applications and what current frameworks fall short of  (Section~\ref{subsec:need_for_framework}).

\subsection{Prompt Engineering}
\label{subsec:prompt_engineering}

Since the introduction of GPT-3, \emph{in-context learning}~\cite{brown2020language} has emerged as a core capability of modern LLMs. By leveraging zero-shot or few-shot examples embedded directly in the input sequence, LLMs adapt to myriad downstream tasks without parameter updates. This phenomenon gave rise to \textit{prompt engineering}~\cite{wei2022chain,shinn2024reflexion}, where the arrangement and content of these context examples can significantly affect model performance. Reported gains of up to 40\% in certain tasks underscore the power of carefully crafted prompts \cite{khattab2022demonstrate}.

Prompt engineering has also expanded beyond simple single-step settings. Retrieval-Augmented Generation (RAG)~\cite{lewis2020retrieval,khattab2022demonstrate} extends LLMs' knowledge by incorporating retrieved evidence to reduce hallucination and improve factual consistency. Meanwhile, \emph{Chain-of-Thought} (CoT) prompting \cite{wei2022chain} instructs the LLM to produce step-by-step reasoning, often boosting interpretability and accuracy. Autonomous agent frameworks such as ReAct \cite{yao2022react} and Reflexion \cite{shinn2024reflexion} further extend LLM usage by incorporating tools and iterative feedback loops. Across these applications—spanning chatbots, multi-hop retrieval, and agentic planning—prompt engineering remains crucial yet increasingly complex.

Despite the growing importance of prompt design, most industry and research efforts currently employ manual engineering. This approach, while intuitive, exhibits several downsides:
\begin{itemize}
    \item \textbf{Time- and labor-Intensive} Each new model or application often demands prompt design from scratch, encumbering both prototyping and deployment cycles.
    \item \textbf{Prompt Sensitivity:} Even small changes in wording can drastically shift model outputs \cite{liu2024lost}, making manual tuning a trial-and-error process with no consistent framework for improvement.
    \item \textbf{Complex Interdependencies:} As LLM pipelines grow to include multiple calls, functional components(such as retriever and user defined data processing functions)~\cite{khattab2022demonstrate,yao2022react}, holistic prompt tuning becomes unmanageable. Graph-like execution flows introduce intricate dependencies that are not easily resolved by hand.
\end{itemize}

\subsection{Automatic Prompt Engineering (APE)}
\label{subsec:ape}

A growing line of research aims to streamline prompt creation and refinement. One approach uses an LLM equipped with a “prompt engineer persona” \cite{yang2023large} or domain-specific heuristics \cite{brown2020language}—often referred to as “meta-prompts” \cite{zhou2022large}—to propose improved textual instructions. Another prominent direction is \emph{Text-Grad} \cite{pryzant2023automatic,yuksekgonul2024textgrad}, which draws a direct analogy to gradient descent by employing a “backward engine” LLM to produce iterative textual feedback that corrects model errors. Concurrent work explores automatic selection and validation of few-shot exemplars, sometimes enhanced by reasoning from a more powerful “teacher” LLM to boost in-context learning \cite{khattab2022demonstrate}, or combines instruction tuning with demonstration (as in DsPy’s MIPRO \cite{opsahl2024optimizing}). DsPy~\footnote{\url{https://github.com/stanfordnlp/dspy}} is by far the only library that have extended these optimization methods into multi-node system, yet it does not offer a fully \emph{auto-differentiable} solution. This limitation prevents the system from pinpointing, for each individual training example, precisely where errors arise in a complex pipeline—missing the greatest advantage of robust \emph{auto prompt engineering} in truly end-to-end LLM-based workflows.



Reinforcement learning has also been applied to prompt engineering: RLPrompt~\cite{deng2022rlprompt} and Tempera~\cite{zhang2022tempera} adjust prompts to maximize a reward signal related to downstream metrics. Some methods pursue ``soft prompting,'' i.e., optimizing a small set of additional embedding tokens \cite{deng2022rlprompt}. While these can be effective, they may suffer from reduced interpretability or limited applicability when gradient access to the LLM is restricted. In addition, although most APE studies regard the LLM weights as fixed, recent work explores blending partial finetuning and prompt optimization. For instance, \cite{soylu2024fine} integrates instruction tuning (or DPO) with few-shot learning to co-optimize model parameters and textual prompts, echoing a broader push to treat hyperparameters, prompts, and model weights as a unified space of adaptation.



Recently, \cite{wang2024correctly} attempted to address multi-node prompt optimization by proposing a \emph{Graph-based Agentic System Optimization (GASO)} framework. Their system uses semantic gradients to propagate textual feedback in a reverse-mode style, sharing conceptual similarities with how our method generalizes Text-Grad to multi-node graphs. Although both works seek to move beyond \emph{single-node} textual gradient methods, \cite{wang2024correctly} focused largely on chain-structured tasks (e.g., \textit{BIG-Bench Hard}, \textit{GSM8K}, and \textit{LIAR}). In contrast, our framework targets a broader class of multi-component or cyclical LLM workflows, via a series of algorithmic innovations such as supporting non-trainable components, time-sequential calls, peer nodes, and selective gradient computation - thereby offering a much more comprehensive and scalable solution for complex agent or RAG pipelines. \cite{wang2024correctly} also did not fully demonstrate how to handle multiple distinct LLM calls in the same graph, which is organically supported in our system.  


\subsection{Why We Need a More Comprehensive Framework}
\label{subsec:need_for_framework}

Despite these advances, a flexible and powerful multi-component auto-optimization architecture, that can allow developers to build effective auto-differentiable feedback graphs to auto-optimize any LLM application,  is yet to emerge. 
Existing methods often do not consider complex workflows containing retrievers or loops. 
Popular LLM libraries such as LangChain\footnote{\url{https://github.com/langchain-ai/langchain}} and LlamaIndex\footnote{\url{https://github.com/run-llama/llama_index}} facilitate the orchestration of multi-step LLM workflows in a graph-like manner but lack the crucial \emph{auto-optimization} capabilities needed for production-scale deployment. While DsPy offers certain auto-optimization features, it primarily focuses on token-heavy few-shot approaches and less effective instruction tuning, sidestepping the central learning paradigm that has propelled modern AI success—automatic differentiation. Consequently, developers often revert to manual, ad hoc prompt engineering, a strategy that grows impractical and error-prone as systems become more complex. A practical analogy is deep neural networks before auto-differentiation frameworks: without a unifying approach to track dependencies and propagate errors end-to-end, sophisticated network designs remained unwieldy.

\emph{LLM-AutoDiff} aims to close this gap by providing an \textit{auto-grad}-like solution for entire LLM applications. In the same way that PyTorch~\cite{paszke2019pytorch} and TensorFlow~\cite{abadi2016tensorflow} transformed neural network development, we systematically address the complexities of multi-node LLM pipelines, including functional components and cyclical loops, under one coherent umbrella. The result is a new paradigm for large-scale LLM deployment: rather than managing prompts in isolation, developers can treat each sub-prompt as a parameter in a broader computation graph, automating both local refinements and global adaptations.

\section{Methods}
\label{sec:method}

In this section, we first lay out the conceptual underpinnings of \emph{Automatic LLM Application Optimization (ALAO)} and its subset, \emph{Automatic Prompt Engineering (APE)} (Section~\ref{sec:alao_ape}). We then introduce how \emph{textual auto-differentiation} is concretely implemented (Section~\ref{sec:interpreting_text_grad}), focusing on the forward/backward passes within a graph-structured LLM system. Finally, we present our two main contributions:

\begin{enumerate}
    \item \textbf{Auto-Differentiation for Any Compound LLM System}: A principled way to handle multi-component and potentially cyclic architectures, described in Section~\ref{sec:llm_diff}.
    \item \textbf{Efficient Training Techniques}: Methods that selectively compute gradients only for error cases, employ two-stage validation, and more, covered in Section~\ref{sec:token_efficient_text_grad}. We also substantially extend the OPRO prompt optimizer in Section~\ref{sec:GDPO}.
\end{enumerate}

Throughout this section, we draw on the AdalFlow library’s formalism for parameter nodes and textual gradients, ensuring mathematical clarity and consistency with the prior literature.

\subsection{ALAO and APE}
\label{sec:alao_ape}

\paragraph{Automatic LLM Application Optimization (ALAO).}
We define \emph{ALAO} as the process of automatically tuning and improving a compound AI system composed of multiple LLM-based modules, each designed for a particular function (e.g., retrieval, classification, generation). This optimization can include altering prompts, hyperparameters, or even partially finetuning the underlying LLMs. Our focus, however, is on prompt-level optimization, which we refer to as \emph{Automatic Prompt Engineering (APE)}.

\paragraph{Automatic Prompt Engineering (APE).}
APE is the subproblem of ALAO concerned solely with optimizing textual prompts across all LLMs in a multi-component workflow. Formally, we model an LLM system as a directed graph
\[
    \mathcal{G} = (\mathcal{N}, \mathcal{E}),
\]
where $\mathcal{N}$ comprises LLMs or other modules (e.g., retrievers), and $\mathcal{E}$ denotes the directed edges capturing dependencies and data flow—possibly with cycles, as in multi-hop retrieval~\cite{khattab2022demonstrate} or ReACT-style agents~\cite{yao2022react}. If $\{\mathcal{M}_i\}$ denotes the set of LLM nodes, then the prompts $\{\mathcal{P}_i\}$ for each $\mathcal{M}_i$ become the \emph{trainable parameters} for APE.


Often, LLM architectures must fulfill multiple objectives (e.g., retrieving high-quality documents while also generating coherent final answers). We partition the system into subtasks $T_1, T_2, \dots$, each with a corresponding loss $\mathcal{L}_{T_i}$ that depends on a subset of prompts $\{\mathcal{P}_{\mathcal{N}_{T_i}}\}$. A joint, multi-task formulation can thus be written as:
\begin{equation}
\label{eq:multi_task_loss}
    \mathcal{L}(\{\mathcal{P}_i\}) \;=\; \bigcup_{i} \mathcal{L}_{T_{i}}\bigl(\mathcal{P}_{\mathcal{N}_{T_{i}}}\bigr),
    \quad
    \{\mathcal{P}_i\}^* = \arg \min_{\{\mathcal{P}_i\}} \bigcup_{i} \mathcal{L}_{T_i}\bigl(\mathcal{P}_{\mathcal{N}_{T_i}}\bigr).
\end{equation}
A typical example is a multi-hop RAG system (Algorithm~\ref{code:multi_hop_rag_cycle}), where one subtask optimizes subquery prompts (via metrics like MRR~\cite{radev2002evaluating}, Recall@k, Precesion@k), and another subtask refines a final generator prompt for a question-answering score. In \textit{AdalFlow}, each node is a \textit{GradComponent}, similar in spirit to PyTorch’s \textit{Module}. Figure~\ref{fig:auto_graph_agent} demonstrates how multi-hop RAG can naturally introduce cycles (e.g., repeated calls to a query generator and retriever).

\begin{algorithm}[t]
    \caption{Pseudo-code for Multi-hop RAG with Cycles}
    \label{code:multi_hop_rag_cycle}
    \begin{algorithmic}[1]
        \State \textbf{Input:} $\textit{Question}$
        \State \textbf{Output:} $\textit{Response}$
        \State \textbf{Procedure:}
        \State $\textit{context} \gets []$
        \For{$i = 1$ to $2$}
            \State $\textit{query} \gets \text{QueryGenerator}(\textit{Question},\, \textit{context})$
            \State $\textit{context} \mathrel{+}= \text{QueryRetriever}(\textit{Question},\, \textit{query})$
        \EndFor
        \State \textbf{Call} $\text{Generator}(\textit{Question},\, \textit{context})$
    \end{algorithmic}
\end{algorithm}

\subsection{Interpreting Textual Auto-Differentiation in AdalFlow}
\label{sec:interpreting_text_grad}

\paragraph{Dynamic Computation Graph $\mathcal{G}_p$.}
Hereinafter, we use the AdalFlow library as an example to describe our algorithm and implementation ideas. AdalFlow adopts an approach analogous to modern deep-learning frameworks, but in a textual domain. A \emph{Parameter} in AdalFlow may represent a prompt (PROMPT), a few-shot example set (DEMOs), an input or output, or a hyperparameter (e.g., top-$k$ for a retriever). Once the system executes a forward pass through the graph $\mathcal{G}$ (potentially unrolling cycles), AdalFlow captures a dynamic, directed acyclic \emph{parameter graph} $\mathcal{G}_p = (\mathcal{N}_p, \mathcal{E}_p)$ tracing how each parameter affects subsequent computations.

The goal of the auto-differentiable computation graph is two folds: (1) trace the intermediate inputs and outputs of each component, and
(2) backpropagate the ``evaluation" signals across the network, identifying the root cause of the errors, pointing the LLM optimizer to the right direction at proposing a new prompt.
Formally, each node $v \in \mathcal{N}_p$ is computed by:
\begin{equation}
\label{eq:forward_pass}
    v \;=\; f_c \bigl(\mathrm{PredecessorsOf}(v)\bigr),
\end{equation}
where $f_c$ is the transformation function (e.g., an LLM invocation or a functional operation), and $\mathrm{PredecessorsOf}(v)$ denotes all incoming parameters to $v$.

\paragraph{Textual Loss Definition.}
A typical LLM system might produce multiple final outputs, each with its own evaluation metric (classification accuracy, BLEU score, etc.). In AdalFlow, we unify these metrics into a textual “loss component.” For example, if $v$ is a final output node, $v_{gt}$ is the ground-truth label (optional if using an LLM judge~\cite{zheng2023judging}), and $s_v$ is the score from an evaluation function $e_v$, then:
\begin{equation}
    s_v = e_v(v,\, v_{gt}), \quad
    \mathcal{L} = \bigl(\text{e}_{\text{desc},v},\, v,\, v_{gt},\, s_v\bigr).
\end{equation}
Here, $\text{e}_{\text{desc},v}$ is a textual description of the loss, and $\mathcal{L}$ can be stored as a \textit{LOSS\_OUTPUT} parameter. AdalFlow uses templating (e.g., Jinja2) to pass these modules to a \emph{backward engine} LLM, which interprets the objective function in natural language.

\paragraph{Forward Pass.}
During the forward pass, the system executes all nodes in topological (or unrolled) order. As it processes each node, it populates $\mathcal{G}_p$ with newly created parameter nodes. This procedure continues until the final outputs are computed, at which point we evaluate the textual or numeric loss.

\paragraph{Backward Pass (Textual Gradients).}
After computing the forward pass and its resulting loss, AdalFlow constructs textual feedback or “gradients” for each node, flowing backward through $\mathcal{G}_p$.

\paragraph{Gradient for the Final Output Node.}
If $v_o$ is the final output node (e.g., an LLM’s response) with only one successor (the loss component), then:
\begin{equation}
\label{eq:final_node_gradient}
    \frac{\partial \mathcal{L}}{\partial v_o}
    \;=\;
    \text{LLM}_{\text{backward}}\bigl(\mathcal{L}\bigr),
\end{equation}
where $\mathcal{L}$ is directly passed to the backward engine. This results in textual gradient on how $v_o$ is matching the ideal response.

\paragraph{Gradients for Intermediate LLM Nodes.}
For an internal node $v$, its textual gradient aggregates feedback from each successor $w \in \mathrm{SuccessorsOf}(v)$:
\begin{equation}
\label{eq:intermediate_grad}
    \frac{\partial \mathcal{L}}{\partial v}
    \;=\;
    \bigcup_{w \in \mathrm{SuccessorsOf}(v)}
    \text{LLM}_{\text{backward}}\bigl(v,\, w,\, \tfrac{\partial \mathcal{L}}{\partial w}\bigr).
\end{equation}
Here, the backward engine uses information about $v$, $w$, and $\frac{\partial \mathcal{L}}{\partial w}$ (the textual gradient of $w$) to craft feedback on how to modify $v$. If the system has only a single final LLM call, the logic is straightforward: the gradient at $v_o$ is broadcast upstream to the current $v$ which will be prompt $\mathcal{P}_i$.

\paragraph{Textual Gradient Descent (TGD).}
Once the backward pass collects all gradients, the system performs a \emph{textual update step}:
\begin{equation}
\label{eq:textual_update_step}
    \mathcal{P}_{v_{\text{new}}}
    \;=\;
    \text{LLM}_{\text{opt}}\Bigl( v, \,\mathrm{GradientContext}(v), \,\tfrac{\partial \mathcal{L}}{\partial v} \Bigr),
\end{equation}
where $\mathrm{GradientContext}(v)$ might include the node’s inputs/outputs and relevant scoring. The \emph{optimizer LLM} thus proposes a new prompt given the textual gradient.

\paragraph{Mini-Batch Training.}
Following \cite{yuksekgonul2024textgrad}, we split data into $(\mathcal{D}_{\mathrm{train}}, \mathcal{D}_{\mathrm{validate}}, \mathcal{D}_{\mathrm{test}})$. Each mini-batch is processed by:
\begin{enumerate}
    \item Forward pass: $\mathcal{B} = \{(x_i,y_i)\}_{i=1}^{B}$ yields predictions $\{\hat{y}_i\}_{i=1}^B$ through $\mathcal{G}_p$.
    \item Backward pass: A textual loss is computed for each sample, producing textual gradients.  
    \item Prompt update: A new prompt is generated if validation on $\mathcal{D}_{\mathrm{validate}}$ improves.
\end{enumerate}
Finally, the test accuracy on $\mathcal{D}_{\mathrm{test}}$ gauges the system’s performance.

\subsection{Auto-Differentiate Any Compound LLM System}
\label{sec:llm_diff}

\begin{figure}[t]
      \centering
      \includegraphics[width=1\linewidth]{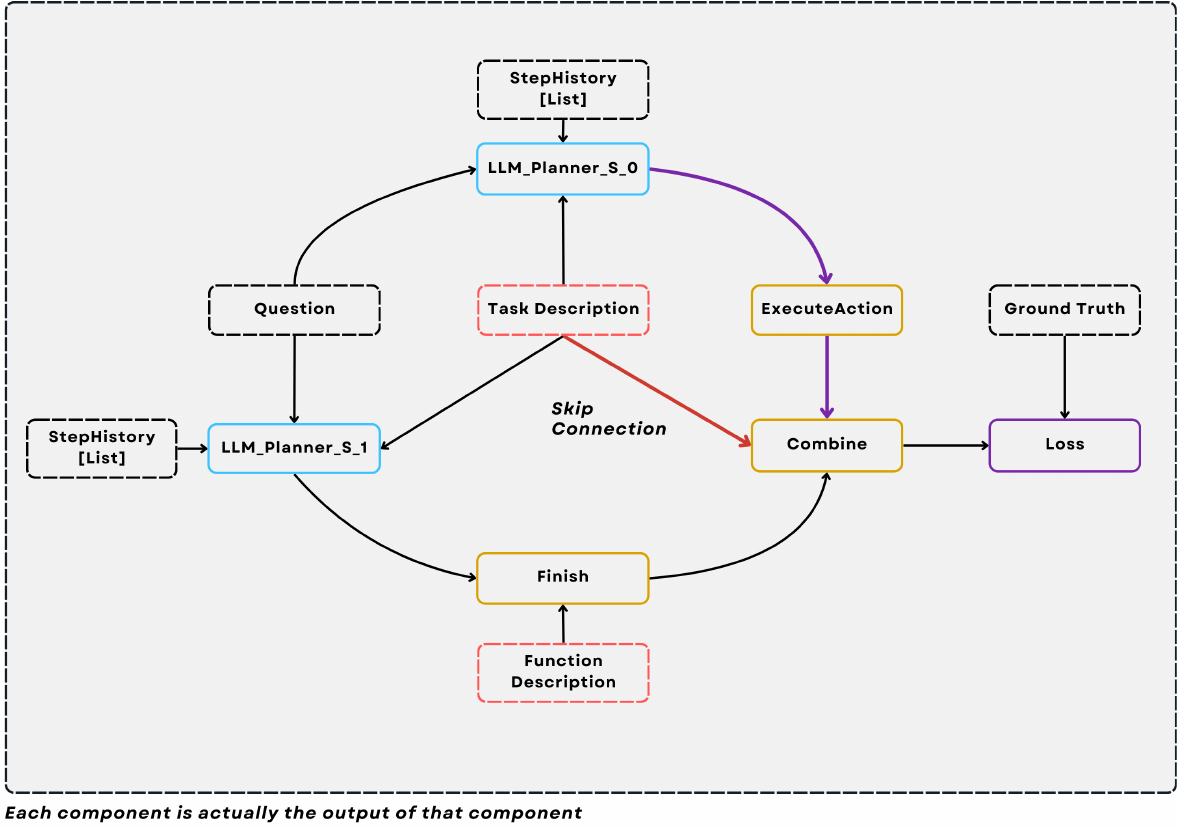}
      \caption{
      \textbf{Example of a ReACT Auto-Differentiable Graph.} 
        This configuration extends our multi-hop RAG paradigm by incorporating a ReACT planner and functional nodes for tool usage and final output assembly. 
        Here, \textit{ExecuteAction} calls the retriever module when needed, while \textit{Finish} is a simple function—governed by its own function docstring—that merges the accumulated step history into a coherent answer. 
        The \textit{Combine} node is a functional component that aggregates outputs from both \textit{ExecuteAction} and \textit{Finish} before producing the final output. 
        Notably, a \emph{skip connection} from the \textit{Task Description} provides direct feedback to the ReACT planner, enabling more targeted textual gradients across the entire system. This design highlights how agent-style loops, functional transformations, and skip connections can all be captured in a unified auto-differentiable framework for LLM-driven workflows.
}
\label{fig:react_auto_graph_agent}

\end{figure}

We now describe how to extend textual auto-differentiation to \emph{any} compound LLM system that may (1) involve multiple LLM calls, (2) interleave functional modules (e.g., retrievers, data-processing functions), and (3) contain cycles through repeated node invocations. Figure \ref{fig:react_auto_graph_agent} illustrates one such example from the practice. These extensions require modifying the original backward-pass formula from Eq.~\eqref{eq:intermediate_grad} to Eq.~\eqref{eq:intermediate_grad_functional} and \eqref{eq:cyclic_grad_call}, ensuring that textual gradients propagate end-to-end in more complex settings.

\subsubsection{Multi-Components}
\label{sec:multi_node_text_grad}

\paragraph{Non-LLM Intermediate Components.}
Modern LLM pipelines often incorporate “intermediate” nodes beyond the LLM calls themselves. For instance, Figure \ref{fig:auto_graph_agent} illustrates two such nodes: 
\begin{enumerate}
    \item \textit{Retriever}, which retrieves documents in response to a sub-query generated by an LLM; 
    \item \textit{CombineList}, which merges and deduplicates two retrieved document lists before feeding them into a final generator node.
\end{enumerate}
Below is a stripped-down Python example for \lstinline|CombineList|, which inherits from \lstinline|GradComponent|:

\begin{lstlisting}[
    basicstyle=\ttfamily\small,
    breaklines=true,
    language=Python,
    keywordstyle=\color{blue}\bfseries,
    stringstyle=\color{orange},
    commentstyle=\color{green!50!black},
    frame=single,
    showstringspaces=false,
    tabsize=4,
    numbers=left,
    numberstyle=\tiny\color{gray}
]
class CombineList(GradComponent):
    def __init__(
        self,
        name="CombineRetrieverOut",
        desc="combines two lists and deduplicate with set"
    ):
        super().__init__(name=name, desc=desc)

    def call(
        self,
        context_1: adal.RetrieverOutput,
        context_2: adal.RetrieverOutput,
        id: str = None
    ) -> List[str]:

        seen = set()
        lists_1 = context_1.documents
        lists_2 = context_2.documents
        combined = [x for x in lists_1 + lists_2 if not (x in seen or seen.add(x))]

        output = adal.RetrieverOutput(
            id=id,
            query=[context_1.query, context_2.query],
            documents=combined,
            doc_indices=[]
        )
        return output
\end{lstlisting}

Because these “functional” nodes have no direct prompt parameters, they do not themselves perform textual updates. However, they still require a backward mechanism to pass gradients upstream when an error arises from \emph{their} output.

\paragraph{Gradients for Intermediate Functional Nodes.}
We categorize these components as \emph{functional nodes} and incorporate two key rules for gradient propagation:
\begin{enumerate}
    \item If a functional node has exactly one predecessor, the gradients on its output simply “pass through” to that predecessor.
    \item If a functional node has multiple predecessors, we must attribute the error signal to the correct predecessor(s). In this case, we adopt \emph{intermediate gradients} that explicitly backpropagate the relevant portion of the feedback to each predecessor.
\end{enumerate}
Formally, for a functional node $w$ with successors \(\mathrm{SuccessorsOf}(v)\), we extend Eq.~\eqref{eq:intermediate_grad} to:
\begin{equation}
\label{eq:intermediate_grad_functional}
    \frac{\partial \mathcal{L}}{\partial v}
    =
    \bigcup_{w \,\in\, \mathrm{SuccessorsOf}(v)}
    \begin{cases}
        \tfrac{\partial \mathcal{L}}{\partial w}, 
        & \text{if $w$ has one predecessor;} \\[6pt]
        \text{LLM}_{\text{backward}}\!\bigl(v,\; w,\; \tfrac{\partial \mathcal{L}}{\partial w}; \text{Template}\bigr),
        & \text{otherwise.}
    \end{cases}
\end{equation}
Here, \(\text{Template}\) refers to a structured text prompt that the backward engine LLM uses to determine how $v$ influenced $w$.

\paragraph{Handling Duplicate Gradients in Multi-Node Systems.}
In more intricate graphs, a single parameter node might receive the same gradient multiple times (e.g., repeated pass-through signals). To prevent unbounded duplication, each gradient is stored in a hash-based data structure keyed by attributes such as the data ID and call index $t$. This approach merges identical feedback, stabilizing the backward pass.

\subsubsection{Cycles}
\label{sec:cyclic_structures}

Certain LLM applications—multi-hop RAG~\cite{khattab2022demonstrate} or iterative ReACT agents~\cite{yao2022react}—may invoke the same node multiple times within a single forward pass. Such cyclical usage accumulates multiple gradients for the same node. 

\paragraph{Time-Sequential Gradients for Cycles.}
To maintain a sensible order of updates, each invocation is indexed by $t$. Let $c$ be a node called $T_c$ times in one pass. We track each call’s gradient:
\[
    \frac{\partial \mathcal{L}}{\partial v^{(t)}} 
    \quad\text{for} \; t \in \{1,\dots,T_c\}.
\]
Then, for trainable parameters $P_c$ feeding into $c$, we gather:
\begin{equation}\label{eq:cyclic_grad_call}
    \frac{\partial \mathcal{L}}{\partial P_c}
    =
    \bigcup_{t=1}^{T_c}
    \Bigl\{\,
    \bigl(t,\;\tfrac{\partial \mathcal{L}}{\partial v^{(t)}}\;\cdot\;\tfrac{\partial v^{(t)}}{\partial P_c}\bigr)
    \Bigr\}.
\end{equation}
Each tuple \(\bigl(t,\dots\bigr)\) preserves the correct temporal sequence. When mini-batches contain multiple data IDs, we also store that ID to avoid mixing gradients across different samples or timesteps.

\vspace{1mm}
\noindent\textbf{Meta-Prompt for Cyclic Structures.}
Our system includes an additional prompt to inform the backward engine about repeated invocations:
\begin{lstlisting}
"If the same DataID has multiple gradients, it means this component/variable is called multiple times in the compound system (with a cycle) 
in the same order they appear in the gradient list."
\end{lstlisting}
This cue helps the LLM-based backward engine handle repeated calls methodically.

\subsubsection{Peers}
\label{sec:peer_nodes}

LLM prompts often mix various “subprompts,” such as instruction text, few-shot examples, and output format. Merging them into a single block can cause:
\begin{enumerate}
    \item \textbf{Noisy or misleading gradients:} It becomes hard to pinpoint which portion of the prompt contributed to an error.
    \item \textbf{Cross-contamination:} An LLM-based optimizer might jumble multiple subprompts, producing incoherent updates.
\end{enumerate}
To address this, we define \emph{peers} as distinct predecessor parameters to a node $v$. For example:
\begin{equation}
\label{eq:peer_nodes}
    P_i = \{\,P_i^{\text{task}},\;P_i^{\text{few-shot}},\;P_i^{\text{out-format}},\dots\}.
\end{equation}
Each subprompt is a separate node in the parameter graph $\mathcal{G}_p$, enabling fine-grained updates.

\paragraph{Peer-Aware Gradients.}
Rather than independently updating each subprompt, we allow a single backward call to produce feedback for \emph{all} peers:
\begin{equation}
\label{eq:peer_grad_generalized}
    \Bigl\{\,
    \tfrac{\partial \mathcal{L}}{\partial u_1},\dots,\tfrac{\partial \mathcal{L}}{\partial u_m}
    \Bigr\}
    =
    \text{LLM}_{\mathrm{backward}}\!\bigl(v,\;\{u_1,\dots,u_m\},\;\tfrac{\partial \mathcal{L}}{\partial v}\bigr),
\end{equation}
where $\{u_1,\dots,u_m\}$ are the subprompt “peer” nodes feeding $v$. This ensures the LLM-based backward engine sees all subprompts jointly, reducing spurious interactions and allowing it to propose updates that are more holistic.

\noindent For completeness, we provide the meta-prompt templates for peer nodes in Appendix~\ref{lst:llm_variables_peers_template}. By recognizing these peer relationships, LLM-Diff avoids confusion in multi-subprompt prompts, resulting in more precise gradient signals and higher-quality proposals.


\subsubsection{Skip Connections as an Option}
\label{sec:skip_connections}

Even with textual auto-differentiation, gradient signals can diminish when LLM-based workflows grow deeper or more branched. This issue parallels vanishing gradients in deep networks, where extra layers compromise end-to-end learning. Drawing inspiration from ResNets \cite{he2016deep} as well as prior attempts to introduce ``training-time" auxilary skips \cite{jaiswal2022training}, we introduce \emph{skip connections} to mitigate gradient attenuation by creating direct feedback pathways that bypass intermediate nodes.

At present, these connections can be flexibly defined by the user. For instance, a node near the pipeline’s end (e.g., a \lstinline|CombineStepHistory| component) can provide a direct route for error signals to flow back to the initial prompt or agent description, in addition to following the standard, “long-path” backward pass. 
Figure~\ref{fig:react_auto_graph_agent} offers a concrete example, where the \textit{Combine} node relays distilled feedback to the \textit{ReACT planner’s} task description. 
Such skip connections help preserve crucial error information that might otherwise vanish as it propagates across multiple LLM or functional nodes. 
Moreover, one can further augment signal strength by injecting intermediate losses or auxiliary objectives at various stages, fostering more stable, fine-grained learning across large LLM applications.

Because skip connections are not automatically created for now, developers remain free to decide when and how to incorporate them based on domain knowledge. AdalFlow currently automates only standard LLM and retriever components; any additional functional node or feedback pathway (including skip connections) must be manually declared. We regard \emph{automated skip-connection discovery} as an exciting avenue for future work, where a meta-learning or heuristic approach could systematically explore which direct feedback routes yield the strongest overall performance.





\subsection{Efficient Training for Textual Auto-Differentiation}
\label{sec:token_efficient_text_grad}

Although textual auto-differentiation offers a flexible mechanism for optimizing prompts, it can be computationally expensive. At each training step, three major operations incur significant cost: 
\begin{enumerate}
    \item \textbf{Backward Pass on Mini-Batches} using a state-of-the-art LLM,
    \item \textbf{Prompt Updates} (i.e., proposals) from an optimizer LLM, and
    \item \textbf{Validation} on the entire validation dataset.
\end{enumerate}
Among these, (1) often dominates as the graph structure gets more complex especially if a large-capacity model (e.g., GPT-4o) is employed for generating textual gradients  and (2) propt proposals follow. Empirical observations from \cite{yuksekgonul2024textgrad} indicate that a single backward pass on a batch of 4 in a single node LLM application can take up to 70 seconds, while generating a single proposal might require 10 seconds. Below, we introduce methods to reduce these overheads while preserving or even improving training effectiveness.

\subsubsection{Compute Gradients Only for Error Samples}
\label{sec:loss_error_samples}

In classical gradient descent, losses from \emph{all} samples accumulate to drive parameter updates. In the textual domain, however, “accumulation” amounts to concatenating loss and gradient information into a prompt for the “optimizer” LLM. This can be unnecessarily costly if many samples are already correct. Indeed, we commonly observe initial accuracies above $80\%$ (e.g., in an object-counting task~\cite{suzgun2022challenging,srivastava2022beyond}), implying that the majority of training tokens can go to correct samples that offer no meaningful feedback on how to fix errors but rather a means for prompt rephrasing. 
Moreover, there is a nontrivial probability that an entire mini-batch of size $n$ contains no error samples at all:
\begin{equation}
\label{eq:chance_no_error}
    P(\text{No errors in batch}) 
    \;=\;
    \frac{\binom{N_1}{n}\,\binom{N_0}{0}}{\binom{N}{n}},
\end{equation}
where $N$ is the dataset size, $N_1 = N \cdot \text{Accuracy}$, and $N_0 = N - N_1$. For instance, at $80\%$ accuracy, $N=50$, and $n=4$, this probability can exceed $40\%$. In such cases, textual gradient prompts would merely rephrase existing prompts rather than supply error-corrective feedback.

To address this inefficiency, we compute gradients only for samples whose performance metric $s(y,\hat{y})$ is below a threshold $\tau$. Concretely, we replace the typical batch loss with the conditional rule:
\begin{equation}
\label{eq:gradient_threshold}
    \frac{\partial \mathcal{L}}{\partial \mathcal{P}}
    \;=\;
    \begin{cases}
        \text{LLM}_{\text{backward}}(\dots, s(y,\hat{y})), & \text{if } s(y,\hat{y}) < \tau,\\[4pt]
        \text{``You score "} + s(y,\hat{y}),               & \text{otherwise}.
    \end{cases}
\end{equation}
High-scoring samples thus bypass the expensive backward pass (receiving only a simple textual note). This strategy both reduces token usage and ensures the backward pass emphasizes cases that truly need improvement. Additionally, it mitigates “lost in the middle” effects~\cite{liu2024lost}, wherein important errors can be diluted within a large block of mostly correct examples.

\subsubsection{Two-Stage Validation}
\label{sec:validation_pass_rate}

Even if we only generate a limited number of new prompts, naïvely validating each candidate on the entire validation set can still be costly. Traditional gradient-based optimization would not require a full validation pass after every single training step. However, in textual optimization, we often must check if the newly proposed prompt truly improves performance. We thus propose a \textbf{two-stage validation}:
\begin{enumerate}
    \item \textbf{Minibatch Validation.} Test the proposed prompt on the current mini-batch $\mathcal{B}_{\text{sub}}$. If accuracy improves, proceed.
    \item \textbf{Full Set Validation.} Evaluate on the full validation set $\mathcal{D}_{\mathrm{val}}$. If the score improves, accept the new prompt; otherwise, revert to the old one.
\end{enumerate}
This hierarchical check prevents the frequent and costly full-dataset evaluations for obviously unhelpful updates. By pruning poor proposals early, we reduce validation overhead without sacrificing reliability.

\paragraph{Multiple Proposals per Backward Pass.}
During each backward pass, we can further optimize efficiency by generating up to $n_p$ candidate prompts in one go and testing them via the two-stage validation. Algorithm~\ref{alg:normal_minibatch} sketches the resulting procedure. If the first candidate fails at the minibatch stage, we simply move to the second one, and so on. This “beam-like” approach helps avoid repeated backward passes when a single candidate is suboptimal.

\begin{algorithm}[t]
    \caption{Token-Efficient Mini-Batch with Two-Stage Validation}
    \label{alg:normal_minibatch}
    \begin{algorithmic}[1]
        \State \textbf{Input:} Graph $\mathcal{G}$, training set $\mathcal{D}_{\mathrm{train}}$, 
        batch size $B$, error threshold $\tau_{\mathrm{correct}}$, max proposals $n_p$, max steps $S_m$.
        \State \textbf{Initialize} prompts and step counter.
        \While{$\mathrm{step} < S_m$}
            \State Sample a mini-batch $\mathcal{B}=\{(x_i,y_i)\}_{i=1}^B$ from $\mathcal{D}_{\mathrm{train}}$.
            \State \textbf{ZeroGrad}: Reset all gradient data structures.
            \State \textbf{Forward}: Compute predictions $\{\hat{y}_i\}_{i=1}^B$ through $\mathcal{G}$.
            \State \textbf{Loss}: For each sample, compute $s_i = e(x_i,y_i,\hat{y}_i)$.
            \State \textbf{Backward}: Based on $s_i$ and threshold $\tau_{\mathrm{correct}}$, compute gradients only for samples with $s_i < \tau_{\mathrm{correct}}$ (Eq.~\ref{eq:gradient_threshold}).
            \For{$k=1$ to $n_p$}
                \State \textbf{Propose}: Generate a new prompt from the optimizer LLM.
                \State Evaluate on $\mathcal{B}$ (\emph{Minibatch Validation}).
                \If{Accuracy improves on $\mathcal{B}$}
                    \State Evaluate on full $\mathcal{D}_{\mathrm{val}}$.
                    \If{Accuracy improves on $\mathcal{D}_{\mathrm{val}}$}
                        \State Accept proposed prompt; \textbf{break}.
                    \Else
                        \State \textbf{Revert}: Discard this proposal.
                    \EndIf
                \Else
                    \State \textbf{Revert}: Discard this proposal.
                \EndIf
            \EndFor
            \If{no proposals accepted}
                \State Retain current prompt and carry $\mathcal{B}$ forward.
            \EndIf
            \State $\mathrm{step} \leftarrow \mathrm{step} + 1$.
        \EndWhile
    \end{algorithmic}
\end{algorithm}

In summary, by selectively computing gradients only for erroneous samples, applying a concise two-stage validation, and exploring multiple prompt candidates at once, we can substantially reduce the computational and token costs of textual auto-differentiation—while still converging on high-quality prompt solutions.

\subsection{Gradient-Driven Prompt Optimizer} 
\label{sec:GDPO}
Beyond textual gradient generation, the prompt optimizer itself plays a pivotal role in determining the final performance of a textual auto-differentiation system. We build on the foundations laid by \textit{Optimization by PROmpting} (OPRO) \cite{yang2023large} and extend its capabilities in three notable ways:
\begin{itemize}
    \item \textbf{Extended Proposal History}. While OPRO stores a record of past prompt-score pairs (e.g., the best-performing prompts), our approach additionally maintains a detailed “proposal history” within each training step. This record (noted as \textit{CH}) encapsulates the specific editing method used (chosen from four possible editing modes) and the rationale behind each prompt modification—along with the standard initial-step history (noted as \textit{SH}). By weaving these local proposals into a global historical context, the optimizer gains deeper insights into how each refinement evolves or fails, thus guiding more targeted updates.
\item \textbf{Peer Awareness for Multi-Prompt Systems}. Modern LLM architectures often combine various subprompts—for instance, instructions, few-shot examples, and output format. Our optimizer explicitly acknowledges this modular structure, allowing it to edit or recompose subprompts (peers) independently while preserving their roles. Such “peer awareness” prevents cross-contamination, ensuring that an improvement to one subprompt does not inadvertently degrade another.
\item \textbf{System Awareness in Multi-Component Pipelines}. When an LLM application involves multiple interdependent components (e.g., retrieval modules feeding into an LLM generator), the optimizer must be cognizant of each component’s function and of how prompts coordinate system-wide. We incorporate “system awareness” by exposing metadata about each prompt’s role, so the optimizer can propose revisions that more effectively harmonize with other prompts or pipeline components.
\end{itemize}
We collectively name our novel enhancements of ORPO as \textit{Gradient-Driven Prompt Optimizer} (\textbf{GDPO}). GDPO allows the gradient-driven optimizer to produce more cohesive and robust prompt updates—whether dealing with single LLM prompts or large, multi-LLM systems with intricate data flows, therefore paving the way for an end-to-end solution that matches the complexity of modern LLM applications.

\begin{figure}[ht]
\centering
\includegraphics[width=\linewidth]{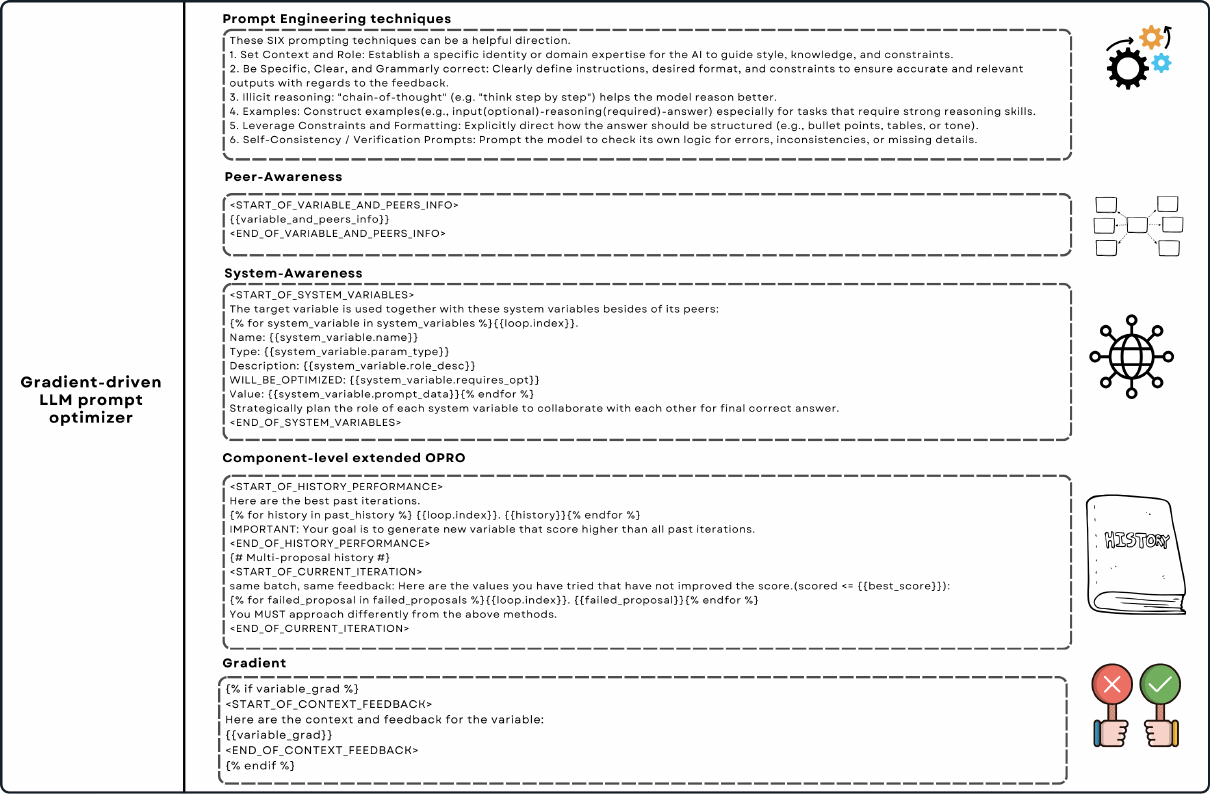}
\caption{\textbf{Gradient-Driven Prompt Optimizer.} The optimizer LLM is prompted with the textual gradients and the previous prompt, along with a multi-node “system view.” It proposes new subprompt texts aligned with the identified errors.}
\label{fig:gradient_driven_optimizer}
\end{figure}

\section{Experiments}
\label{headings}

In this section, we evaluate \textbf{LLM-AutoDiff} across five types of LLM pipelines, each designed to probe different aspects of multi-node and multi-hop workflows. We begin by outlining our experimental setup (Section~\ref{sec:exp_setup}), then discuss results (Section~\ref{sec:exp_results}), and finally present ablation studies to isolate the roles of textual gradients, prompt engineering techniques, and historical information in our optimizer (Section~\ref{sec:ablation_studies}).

\subsection{Experimental Setup}
\label{sec:exp_setup}

\paragraph{Pipelines Overview.}

We design five pipelines, ranging from a single LLM performing simple classification to multi-LLM or agentic architectures on HotPotQA~\cite{yang2018hotpotqa}.

\begin{itemize}
    \item \textbf{One-LLM Workflow.} A single LLM answers a consistent style of question:
          \begin{enumerate}
              \item \textit{ObjectCount}~\cite{srivastava2022beyond,suzgun2022challenging}: We randomly split 50/100/100 samples for training/validation/test from the original 500-sample subset, evaluating with an Exact Match (EM) metric.
              \item \textit{TREC-10}~\cite{li-roth-2002-learning}: 120/166/344 samples for train/val/test (sub-sampled from the original test set). We also use EM here, following \cite{yang2018hotpotqa}.
          \end{enumerate}

    \item \textbf{Multi-Node RAG Workflows.} On the HotPotQA dataset, we construct four variations of retrieval-augmented generation (RAG) pipelines:
          \begin{enumerate}
              \item \textit{Vanilla RAG}: 2-component system (retriever + generator),
              \item \textit{Multi-hop RAG}: 4 components (two subquery generators, one retriever, one final generator),
              \item \textit{Multi-hop RAG (Cycle)}: 3 components (one LLM subquery generator called twice, a retriever, and a final generator),
              \item \textit{Agentic RAG}: A ReAct-style agent~\cite{yao2022react} with two tools (\texttt{Retriever} and \texttt{Finish}) and up to three steps. 
          \end{enumerate}
          We follow DsPy's standard top-$3$ retrieval for Vanilla RAG and top-$2$ for the others. For training, we sample 50 “hard” queries from the official HotPotQA training set and split the 7405-test subset into a 100-sample validation set and a 200-sample final test set.

For certain comparisons, we also reference DsPy’s multi-node capabilities \cite{khattab2022demonstrate}, but note that its agentic RAG differs considerably, so we primarily cite reported metrics. Additional implementation details (e.g., model hyperparameters,code) are in Appendix~\ref{appendix:experimental_implementation}.
\end{itemize}

\paragraph{Baselines and Comparisons.}
We denote:
\begin{itemize}
    \item \textbf{TG}: Text-Grad \cite{yuksekgonul2024textgrad},
    \item \textbf{DsPy}: The multi-node auto-prompt library from \cite{opsahl2024optimizing}, specifically using \textit{MIPROv2} (Bayesian instruction tuning + few-shot learning),
    \item \textbf{Ours}: \textbf{LLM-AutoDiff} as presented in this paper.
\end{itemize}
We adopt the \textit{DsPy(n+m)} notation to denote $n$ demonstration samples plus $m$ raw samples used. Where relevant, \textit{DsPy$^*$} refers to reported numbers from their original paper (which may use slightly different dataset splits). All methods are evaluated with the same EM or $F_1$ criteria, implemented as \texttt{EvalFn} in AdalFlow.

For \textit{Ours}, we conduct 12 training steps at a batch size of 4, mirroring \textit{TG}’s standard setting. We choose the best checkpoint based on validation accuracy before reporting final performance on the test set. Our MIPROv2 baseline (DsPy) is configured as follows:
\begin{lstlisting}
tp = MIPROv2(
    metric=validate_answer,
    prompt_model=gpt_4,
    task_model=turbo,
    num_candidates=12,
    init_temperature=1.0
)
\end{lstlisting}

\paragraph{Implementation Details.}
\begin{itemize}
    \item We use \texttt{GPT-3.5-turbo-0125} as the forward engine (i.e., the main LLM performing tasks).
    \item \texttt{GPT-4o-2024-08-16} is used as the frozen “backward engine” and “optimizer” LLMs.
    \item On HotPotQA, we use $F_1$ \cite{yang2018hotpotqa} rather than strict EM for the internal loss, as partial matches better reflect incremental improvements during minibatch validation.
    \item For each experiment, we measure starting accuracy and the final trained accuracy on both the validation and test set. Hyperparameters and additional code details can be found in Appendix~\ref{appendix:experimental_implementation}.
\end{itemize}

Table~\ref{tab:tasks_datasets} summarizes each dataset and pipeline.





\subsection{Results}
\label{sec:exp_results}

Table~\ref{tab:llm_task_pipeline} presents an overview of validation and test accuracies across our five pipelines. We highlight a few key observations:

\begin{enumerate}
    \item \textbf{Single LLM Pipelines.} 
    On \textit{ObjectCount}, \textit{TG} reported a final $91.9\%$ in prior work, but in our replication with a new split, we see a lower $84.5\%$ test accuracy after 4 runs (despite a high $95.25\%$ on validation). In contrast, our method (\textit{Ours}) achieves $93.75\%$ on test with less overfitting, outperforming \textit{TG} by $9.25\%$. Similarly, on \textit{TREC-10}, \textit{Ours} is $4.2\%$ above \textit{TG} and $7.9\%$ above DsPy, while using fewer tokens.
    
    \item \textbf{HotPotQA RAG Pipelines.}
    We evaluated four distinct RAG workflows, achieving an average performance improvement of \textbf{10\%} across tasks. The most notable gain occurred with \textit{Agentic RAG}, where performance doubled after only 12 training steps, starting from \textbf{16.5\%} test accuracy with default prompts and tool descriptions.  Notably, the \textit{Agentic RAG} pipeline—though the most flexible—yields the lowest absolute accuracy across the board (consistent with \cite{opsahl2024optimizing}). We suspect this is due to the ReAct planner’s multi-task load: it must decide retrieval steps, interpret partial results, and generate final answers, often leading to longer prompts and higher complexity. Even so, our gradient-driven approach outperforms zero-shot or pure few-shot baselines in nearly every RAG variant. For example, on \textit{Vanilla RAG} and \textit{Multi-hop RAG (Cycle)}, \textit{Ours} exceeds \textit{DsPy(2+2)} in test accuracy by a comfortable margin, underscoring the benefits of multi-node textual gradient propagation.

    \item \textbf{Efficiency Gains.}
    Besides higher accuracy, our method is notably faster and more token-efficient. On \textit{ObjectCount}, we consume fewer tokens over 12 steps than \textit{TG} does in exploring just 12 proposals. Empirically, \textit{Ours} also converges in less wall-clock time thanks to techniques like selective gradient computation and two-stage validation (Section~\ref{sec:token_efficient_text_grad}).
\end{enumerate}

Overall, \textbf{LLM-AutoDiff} consistently outperforms the standard Text-Grad method in both single-node and multi-node RAG scenarios, while significantly reducing training overhead. Although DsPy offers competitive results in certain multi-hop pipelines, it relies heavily on few-shot sampling and lacks a true, node-level textual backpropagation mechanism. By contrast, our gradient-driven approach not only pinpoints the sources of error in more complex workflows but also leverages intermediate input-output traces to generate on-the-fly demonstrations—arguably a more flexible and token-efficient strategy than DsPy’s rejection-based sampling \cite{khattab2022demonstrate, opsahl2024optimizing}. This result suggests that local, gradient-informed “instruction tuning” can be more effective than methods in which the “optimizer” only sees final-task data and prompts. We provide further evidence for these advantages in the ablation studies below (Section~\ref{sec:ablation_studies}).

\begin{table*}[ht!]
    \centering
    \fontsize{8.5}{10}\selectfont

    \begin{tabularx}{\textwidth}{>{\hsize=0.15\hsize}X >{\hsize=0.15\hsize}X >{\hsize=0.2\hsize}X >{\hsize=0.35\hsize}X >{\hsize=0.15\hsize}X}

        \toprule
        \textbf{LLM Task Pipeline} & \textbf{Dataset}          & \textbf{Task Type}            & \textbf{Components}                                                & \textbf{LLM Calls}  \\
        \midrule
        \multirow{3}{*}{One LLM  } & ObjectCount               & Simple QA on counting objects of a particular category                     & 1 (generator)                                                      & 1                   \\
        \cmidrule{2-5}
                                   & TREC-10                    &  Classify a question into one of 6 coarse classes               & 1 (generator)                                                      & 1                   \\
        \midrule
        Vanilla RAG                & \multirow{4}{*}{HotPotQA} & \multirow{4}{*}{Multi-hop QA} & 2 (retriever + generator)                                          & 1                  \\
         \cmidrule{4-5}
        Multi-hop RAG              &                           &                               & 4 (query\_generator\_0, query\_generator\_2, retriever, generator) & 3                   \\
        \cmidrule{4-5}
        Multi-hop RAG(Cycle)       &                           &                               & 3  (query\_generator, retriever, generator)                        & 2                   \\
        \cmidrule{4-5}
        Agentic RAG                &                           &                               & 3 (ReAct, retriever as a tool, finish as a tool)                   & 4 (max 4 steps) \\

        \bottomrule
    \end{tabularx}
    \caption{Summary of our tasks, datasets and pipelines for benchmarking.}
    \label{tab:tasks_datasets}
\end{table*}

\begin{table*}[ht!]
    \centering

    \fontsize{8.5}{10}\selectfont

    \begin{tabularx}{\textwidth}{>{\raggedright\arraybackslash}p{0.15\textwidth} >{\raggedright\arraybackslash}p{0.1\textwidth} >{\raggedright\arraybackslash}p{0.07\textwidth} >{\raggedright\arraybackslash}p{0.1\textwidth} >{\raggedright\arraybackslash}p{0.13\textwidth} C{0.15} >{\raggedright\arraybackslash}p{0.15\textwidth} >{\raggedright\arraybackslash}p{0.06\textwidth}}

        \toprule
        \textbf{LLM Task Pipeline} & \textbf{Dataset}              & \textbf{Metric}           & \textbf{Method} & \textbf{Start Acc(\%)(v, t)} & \textbf{Val Acc (\%)}         & \textbf{Test Acc}(\%)         \\
        \midrule

        \multirow{9}{*}{One LLM}
                                   & \multirow{3}{*}{ObjectCount}  & \multirow{3}{*}{EM}       & \textbf{Ours}   & \textbf{86, 83}              & \textbf{96.5 $\pm$ 2.68(98) } & \textbf{93.75 $\pm$ 1.12(95)} \\ 
                                   &                               &                           & TG              & {74, 78}                     & $95.25 \pm 1.48$(97)          & $84.5 \pm 3.6$(89)            \\
                                   &                               &                           & DsPy(5+2)       & {62, 60}                     & $84 \pm 4.47$                 & $82.5 \pm 5.2$                \\
        \cmidrule{2-7}
                                   & \multirow{3}{=}{TREC-10}       & \multirow{3}{*}{EM}       & \textbf{Ours}   & 78.9, 81.7                   & \textbf{86.9$\pm$4.09(92.16)} & \textbf{87.5$\pm$3.94(91)}    \\                                              
                                   &                               &                           & TG              & \textbf{80.12,83.53}         & $83.73\pm1.13(85)$            & $84.88\pm1.29$(86.8)          \\
                                   &                               &                           & DsPy(5+2)       & 72.89, 76.35                 & $86.4\pm0.74(87.4)$           & $81.7\pm1.23$(83.1)           \\
   
        \midrule
        \multirow{2}{*}{Vanilla RAG}
                                   & \multirow{2}{*}{hotpot\_qa}   & \multirow{2}{*}{EM}       & \textbf{Ours}   & \textbf{37, 39.5}            & \textbf{47$\pm$1.22(49)}      & \textbf{43.25$\pm$2.66(46)}   \\
                                   &                               &                           & DsPy(0)         & \multirow{2}{*}{25, 29.5}    & $26.25\pm2.86(30)$            & $29.75\pm2.19(31.5)$          \\
                                   &                               &                           & DsPy(4)         &                              & $40.5\pm2.29(43)$             & $42.375\pm3.78(47.5)$         \\

        \midrule

        \multirow{2}{*}{Multi-hop RAG}
                                   & \multirow{2}{*}{hotpot\_qa}   & \multirow{2}{*}{EM}       & \textbf{Ours}   & \textbf{48, 40.5}            & \textbf{56.25 $\pm$ 1.92(59)} & {$48.25\pm0.75$(49.5)}        \\
                                   &                               &                           & DsPy(0)         & \multirow{2}{*}{32, 36.5}    & $35.5\pm4.33(42)$             & $37.875\pm5.14(46)$           \\
                                   &                               &                           & DsPy(2+2)       &                              & $47.25\pm 3.03$ (50)          & \textbf{50.63$\pm$3.08(53)}   \\
        \midrule
        \multirow{2}{=}{Multi-hop RAG(Cycle)}
                                   & \multirow{2}{*}{hotpot\_qa}   & \multirow{2}{*}{EM}       & \textbf{Ours}   & \textbf{48, 40.5 }           & \textbf{58$\pm$0.7(59) }      & \textbf{49.625$\pm$2.9(53)}   \\
                                   &                               &                           & DsPy(0)         & \multirow{2}{*}{32, 36.5}    & $41.25\pm 4.82(45)$           & $37.875\pm 3.64(42.5)$        \\
                                   &                               &                           & DsPy(2+2)       &                              & $46.76\pm5.12(53)$            & $47.75\pm0.75(48.5)$          \\

        \midrule

        \multirow{2}{=}{Agentic RAG }
                                   & \multirow{2}{*}{hotpot\_qa}   & \multirow{2}{*}{EM}       & \textbf{Ours}   & 15, 16.5                       & $35.5\pm 2.06$(38)           & \textbf{32.25$\pm$ 2.46(35.5)}     \\
                                   &                               &                           & DsPy$^*$        & N/A                          & N/A                           & 31$^*$                        \\

        \bottomrule
    \end{tabularx}
    \caption{Comparison of validation/test accuracies for different LLM task pipelines. ``Start Acc'' is the initial (zero-step) accuracy before auto-optimization. Parentheses show the highest observed accuracy within 12 steps. Agentic RAG is the most flexible but also yields the lowest absolute accuracy. Our method consistently outperforms baselines on single-node tasks and several multi-node RAG workflows while using fewer tokens and converging faster. Note that \textit{DsPy$^*$} refers to reported numbers from their original paper (which may use slightly different dataset splits).}
    \label{tab:llm_task_pipeline}
\end{table*}

\subsection{Ablation Studies}
\label{sec:ablation_studies}

To disentangle the roles of textual gradients, prompt-engineering heuristics, and historical memory in the optimizer, we conduct a series of ablation experiments on selected tasks (ObjectCount, TREC-10, and HotPotQA pipelines). We compare:
\begin{enumerate}
    \item \textit{With vs.\ without gradients}: If gradients are disabled, we rely on raw input-output-score pairs (akin to OPRO~\cite{yang2023large}).
    \item \textit{With vs.\ without prompt engineering techniques}: We remove common prompt-engineering injections from the meta-prompts fed to the optimizer.
    \item \textit{With vs.\ without history}: Some variants see multiple past proposals (e.g., \textit{OPRO(w.o data)(5)}) while others limit the history to 2 or 5 prompts.
\end{enumerate}

\paragraph{Results.}
Table~\ref{tab:ablations} shows that textual gradients deliver the largest gains in more complex pipelines, confirming that multi-component workflows benefit most from a fully auto-differentiable approach. Prompt engineering inside the optimizer also contributes sizable improvements on single-LLM tasks, and historical memory ensures stable convergence by letting the optimizer learn which edits have or have not been successful in previous steps. From Table~\ref{tab:ablations}, we conclude:
\begin{itemize}
    \item \textbf{Gradients Matter Most for Complex Pipelines.} Removing textual gradients (i.e., OPRO with data but no node-level backprop) degrades performance considerably in multi-hop or multi-node tasks. In simpler single-LLM scenarios, the drop is smaller because the optimizer can still correlate input-output scores with final accuracy. For multiple LLM nodes, however, localized textual gradients become essential to pinpoint each node’s error contribution.
    \item \textbf{Prompt-Engineering Aids Single-Node Tasks.} Disabling common prompt-engineering cues in the optimizer (``w/o PE'') leads to noticeable performance drops in simpler tasks like TREC-10 and ObjectCount, where standard engineering heuristics (e.g., clarifying question type) strongly boost accuracy.
    \item \textbf{History Improves Stability.} Maintaining a rolling record of best prompts (\textit{SH} + \textit{CH}) consistently outperforms variants with minimal or no history. The LLM optimizer can learn from previously accepted or rejected edits, preventing re-introduction of poor patterns.
\end{itemize}

Collectively, these results support the core idea that textual gradients with node-level feedback are crucial for multi-component systems, while injection of well-known prompt-engineering heuristics and an informed history further stabilize and enhance performance.

\begin{table}[ht!]
    \centering
    \fontsize{9}{11}\selectfont
    \begin{tabularx}{\textwidth}{>{\raggedright\arraybackslash}p{0.22\textwidth}>{\raggedright\arraybackslash}p{0.15\textwidth} >{\raggedright\arraybackslash}p{0.15\textwidth} >{\raggedright\arraybackslash}p{0.17\textwidth} >{\raggedright\arraybackslash}p{0.1\textwidth} >{\raggedright\arraybackslash}p{0.1\textwidth} }
        \toprule
        \textbf{Task Pipeline}                        & \textbf{Method}   & \textbf{Validation Accuracy (\%)}  & \textbf{Test Accuracy (\%)}       & \textbf{Minibatch Pass Rate (\%)} & \textbf{Validation Pass Rate (\%)} \\
        \midrule
        \multirow{7}{*}{One LLM (ObjectCount)}        & Ours(2)           & $96.5 \pm 2.68$                    & $93.75 \pm 1.12(95)$              & 58.6                              & 12.8                               \\ 
                                                      & OPRO(w.o data)(2) & $92.5\pm2.87$ \textit{(-4)}        & $84.75\pm1.92$ \textit{(-9)}      & 56.7                              & 10                                 \\
                                                      & OPRO(w.o data)(5) & $92.5\pm1.12$  \textit{(-4)}       & $86.5\pm2.06$ \textit{(-7.2)}     & 59                                & 9.4                                \\
        \cmidrule{2-6}
                                                      & OPRO(w data)(2)   & $95.5\pm2.05$   \textit{(-1)}      & $92.75\pm2.59(96)$  \textit{(-1)} & 71.6                              & 13.1                               \\
        \cmidrule{2-6}
                                                      & Ours(w/o PE)      & $95 \pm 1.87$      \textit{(-1.5)} & $86 \pm 4.74$  \textit{(-7.75)}   & 77.5                              & 10.5                               \\
                                                      & Ours(w/o H)       & $96.5 \pm 1.11$                    & $94 \pm 1.58(96)$                 & 79.6                              & 11.8                               \\

        \midrule
        \multirow{6}{*}{One LLM (TREC-10) }            & Ours(2)           & $86.9 \pm 4.09$                    & $87.5 \pm 3.94$                   & 76.7                              & 8.8                                \\
                                                      & OPRO(w.o data)(2) & $82.9 \pm 1.83$ \textit{(-4)}      & $81.3\pm1.76$  \textit{(-6.2)}    & 15.3                              & 10.4                               \\
                                                      & OPRO(w.o data)(5) & $85.1\pm3.2$\textit{(-1.8)}        & $85.6\pm 2.44$ \textit{(-1.9)}    & 26.1                              & 17.3                               \\
     \cmidrule{2-6}
                                                      & OPRO(w data)(2)   & $84.2\pm3.4$ \textit{(-2.7)}       & $85.1\pm 2.1$  \textit{(-2.4)}    & 36.6                              & 14.4                               \\
      \cmidrule{2-6}
                                                      & Ours(w/o PE)      & $83 \pm 1.4$ \textit{(-3.9)}       & $85.25 \pm 2.2$ \textit{(-2.25)}  & 36.4                              & 15.4                               \\
                                                      & Ours(w/o H)       & $82.5 \pm 2.79$   \textit{(-4.4)}  & $83.6 \pm 3.11$  \textit{(-3.9)}  & 26.1                              & 12.2                               \\

        \midrule
        \multirow{2}{*}{ Vanilla RAG (hotpot\_qa) }   & Ours(2)           & $47 \pm 1.22$                      & $43.25 \pm 2.66$                  & 26.9                                & 18.4                                 \\

                                                      & OPRO(w.o data)(2) & $43 \pm 1.22$  \textit{(-4)}       & $ 40.5\pm2.15$  \textit{(-2.75)}  & 22.1                              & 15.5                               \\
                                                         & OPRO(w.o data)(5)   & $40.25\pm1.5$  \textit{(-6.75)}    & $41.1\pm2$ \textit{(-2.15)}    & 18.5                            & 15.8                           \\
     \cmidrule{2-6}                                                    
                                                        & OPRO(w data)(2)   & $43.25\pm0.8$  \textit{(-3.75)}    & $40.75\pm2.02$ \textit{(-2.5)}    & 24.75                            & 14.9                                  \\
                                                     
       \cmidrule{2-6}                                            
                                                      & Ours(w/o PE)      &  $ 45.24\pm0.4$  \textit{(-1.76)}                                  & $ 42.4\pm1.43$  \textit{(-0.85)}                                  & 25.2                                  &  26.1                                  \\
                                                      & Ours(w/o H)       & $43.25 \pm 1.64$  \textit{(-3.75)} & $42.4\pm1.88$ \textit{(-0.85)}    & 18.1                              & 37.7                               \\
        \midrule

        \multirow{3}{*}{ Multi-hop RAG (hotpot\_qa) } & Ours(2)           & $56.25 \pm 1.92$                   & $48.25 \pm 0.75$                  &28.3                         & 21.7                                 \\
                                                      & OPRO(w.o data)(2) & $51.75\pm2.28$ \textit{(-4.5)}     & $43.87\pm2.16$\textit{(-4.38)}    & 13.4                              & 18.5                               \\
        \cmidrule{2-6}
                                                      & OPRO(w data)(2)   & $51.8\pm1.09$  \textit{(-4.45)}    & $46.4\pm2.53$ \textit{(-1.85)}    & 24.2                              & 11                                 \\
                \cmidrule{2-6}
                                              
                                                       & Ours(w/o PE)       & $54.25\pm2.95$    \textit{(-2)}    & $48.1\pm2.43$ \textit{(-0.15)}      & 49.2                              & 10.5                               \\
                                                      & Ours(w/o H)       & $52.25\pm3.27$    \textit{(-4)}    & $47\pm2.76$ \textit{(-1.25)}      & 23.7                              & 13.3                               \\

        \bottomrule
    \end{tabularx}
\caption{Ablation on textual gradients, prompt engineering, and history. ``Minibatch Pass Rate'' denotes the percentage of proposed prompts that pass minibatch validation; ``Validation Pass Rate'' denotes the percentage that also pass the full validation set.}
    \label{tab:ablations}
\end{table}

\section{Conclusion and Future Work}
\label{sec:conclusion}

In this work, we have shown that textual backpropagation can be extended into a full-fledged auto-differentiation paradigm for complex LLM applications, encompassing multi-component pipelines, functional intermediate nodes, and even cyclical invocations. Our framework unifies the idea of gradient-like updates with the flexibility of natural language, allowing prompts at every node---whether they be few-shot examples, task instructions, or output formatting---to be iteratively improved via systematic error feedback. By coupling these textual gradients with a sophisticated optimizer and cost-saving strategies (e.g., focusing on error cases, two-stage validation), we demonstrate that large-scale prompt engineering can be automated in a manner analogous to how deep learning libraries handle numeric gradients.

Despite these advances, our work represents only the first step toward \emph{Automatic LLM Application Optimization (ALAO)} in truly general settings. Modern AI pipelines frequently span beyond prompts, integrating finetuned model components, hyperparameter searches (e.g., for chunk sizes in RAG), and specialized data transformations. Moving forward, several promising research directions arise:
\begin{itemize}
    \item \textbf{Hyperparameter and Architectural Co-Optimization.}
While our current focus is on textual prompts, many real-world systems intertwine prompt decisions with hyperparameters (retrieval depth, chunk lengths, beam sizes) or even partial finetuning (SFT, LoRA). Extending our framework to unify textual gradient feedback with model- or hyperparameter-level changes could yield more holistic improvements.
\item \textbf{Advanced Validation and Data Labeling.}
As with traditional machine learning, the quality and quantity of labeled data strongly constrain what can be optimized. Future work could explore adaptive data labeling within the AdalFlow library, automatically identifying or acquiring the most valuable training samples, or leveraging LLMs-as-judges~\cite{zheng2023judging} to seamlessly integrate unlabeled data.
\item \textbf{Dynamic and Reconfigurable Graphs.}
Many agentic pipelines undergo structural changes over time, adding or removing modules (e.g., new retrieval tools) or updating chain-of-thought logic. Building a dynamically \emph{reconfigurable} version of the graph would allow textual gradients to be consistently applied even as the system’s topology evolves.
\item \textbf{Deeper Integration with Model Finetuning.}
Although textual prompts provide a powerful mechanism for in-context adaptation, it is sometimes beneficial to combine them with partial or full finetuning of the base LLM. Investigating how textual gradients interact with numeric gradients (e.g., parameter-efficient fine-tuning or reinforcement learning) may lead to hybrid ``neurosymbolic" methods that maximize both interpretability (through text) and representational capacity (through model updates).
\item \textbf{Generalization to Other Modalities.}
An intriguing frontier is applying similar “textual backpropagation” principles to multimodal tasks (e.g., image retrieval, video summarization) or code-centric pipelines. In such scenarios, carefully designed textual feedback could still drive updates, but the domain constraints and data flows become more complex.
\end{itemize}
In short, our results underscore that a well-crafted combination of dynamic graphs, textual gradients, and peer-aware prompts can substantially automate prompt engineering in diverse LLM systems. We hope this framework will spur a wave of new research and practical extensions—ranging from deeper integration with model weights to advanced multi-task prompting within AdalFlow. Ultimately, as LLM applications continue to expand in scale and complexity, automatic optimization of both prompts and broader system components will become increasingly indispensable. We invite researchers and practitioners to build upon the groundwork laid here, further refining and extending the capabilities of \emph{Automatic LLM Application Optimization (ALAO)} for the next generation of intelligent systems.

{\small
    \bibliographystyle{ieee_fullname}
\bibliography{main}

\begin{thebibliography}{10}\itemsep=-1pt

\bibitem{abadi2016tensorflow}
Mart{\'\i}n Abadi, Paul Barham, Jianmin Chen, Zhifeng Chen, Andy Davis, Jeffrey Dean, Matthieu Devin, Sanjay Ghemawat, Geoffrey Irving, Michael Isard, et~al.
\newblock $\{$TensorFlow$\}$: a system for $\{$Large-Scale$\}$ machine learning.
\newblock In {\em 12th USENIX symposium on operating systems design and implementation (OSDI 16)}, pages 265--283, 2016.

\bibitem{brown2020language}
Tom~B Brown.
\newblock Language models are few-shot learners.
\newblock {\em arXiv preprint arXiv:2005.14165}, 2020.

\bibitem{deng2022rlprompt}
Mingkai Deng, Jianyu Wang, Cheng-Ping Hsieh, Yihan Wang, Han Guo, Tianmin Shu, Meng Song, Eric~P Xing, and Zhiting Hu.
\newblock Rlprompt: Optimizing discrete text prompts with reinforcement learning.
\newblock {\em arXiv preprint arXiv:2205.12548}, 2022.

\bibitem{he2016deep}
Kaiming He, Xiangyu Zhang, Shaoqing Ren, and Jian Sun.
\newblock Deep residual learning for image recognition.
\newblock In {\em Proceedings of the IEEE conference on computer vision and pattern recognition}, pages 770--778, 2016.

\bibitem{jaiswal2022training}
Ajay~Kumar Jaiswal, Haoyu Ma, Tianlong Chen, Ying Ding, and Zhangyang Wang.
\newblock Training your sparse neural network better with any mask.
\newblock In {\em International Conference on Machine Learning}, pages 9833--9844. PMLR, 2022.

\bibitem{khattab2022demonstrate}
Omar Khattab, Keshav Santhanam, Xiang~Lisa Li, David Hall, Percy Liang, Christopher Potts, and Matei Zaharia.
\newblock Demonstrate-search-predict: Composing retrieval and language models for knowledge-intensive nlp.
\newblock {\em arXiv preprint arXiv:2212.14024}, 2022.

\bibitem{kimllm}
Sehoon Kim, Suhong Moon, Ryan Tabrizi, Nicholas Lee, Michael~W Mahoney, Kurt Keutzer, and Amir Gholami.
\newblock An llm compiler for parallel function calling.
\newblock In {\em Forty-first International Conference on Machine Learning}.

\bibitem{lewis2020retrieval}
Patrick Lewis, Ethan Perez, Aleksandra Piktus, Fabio Petroni, Vladimir Karpukhin, Naman Goyal, Heinrich K{\"u}ttler, Mike Lewis, Wen-tau Yih, Tim Rockt{\"a}schel, et~al.
\newblock Retrieval-augmented generation for knowledge-intensive nlp tasks.
\newblock {\em Advances in Neural Information Processing Systems}, 33:9459--9474, 2020.

\bibitem{li-roth-2002-learning}
Xin Li and Dan Roth.
\newblock Learning question classifiers.
\newblock In {\em {COLING} 2002: The 19th International Conference on Computational Linguistics}, 2002.

\bibitem{liu2024lost}
Nelson~F Liu, Kevin Lin, John Hewitt, Ashwin Paranjape, Michele Bevilacqua, Fabio Petroni, and Percy Liang.
\newblock Lost in the middle: How language models use long contexts.
\newblock {\em Transactions of the Association for Computational Linguistics}, 12:157--173, 2024.

\bibitem{opsahl2024optimizing}
Krista Opsahl-Ong, Michael~J Ryan, Josh Purtell, David Broman, Christopher Potts, Matei Zaharia, and Omar Khattab.
\newblock Optimizing instructions and demonstrations for multi-stage language model programs.
\newblock {\em arXiv preprint arXiv:2406.11695}, 2024.

\bibitem{paszke2019pytorch}
Adam Paszke, Sam Gross, Francisco Massa, Adam Lerer, James Bradbury, Gregory Chanan, Trevor Killeen, Zeming Lin, Natalia Gimelshein, Luca Antiga, et~al.
\newblock Pytorch: An imperative style, high-performance deep learning library.
\newblock {\em Advances in neural information processing systems}, 32, 2019.

\bibitem{pryzant2023automatic}
Reid Pryzant, Dan Iter, Jerry Li, Yin~Tat Lee, Chenguang Zhu, and Michael Zeng.
\newblock Automatic prompt optimization with" gradient descent" and beam search.
\newblock {\em arXiv preprint arXiv:2305.03495}, 2023.

\bibitem{radev2002evaluating}
Dragomir~R Radev, Hong Qi, Harris Wu, and Weiguo Fan.
\newblock Evaluating web-based question answering systems.
\newblock In {\em LREC}. Citeseer, 2002.

\bibitem{shinn2024reflexion}
Noah Shinn, Federico Cassano, Ashwin Gopinath, Karthik Narasimhan, and Shunyu Yao.
\newblock Reflexion: Language agents with verbal reinforcement learning.
\newblock {\em Advances in Neural Information Processing Systems}, 36, 2024.

\bibitem{soylu2024fine}
Dilara Soylu, Christopher Potts, and Omar Khattab.
\newblock Fine-tuning and prompt optimization: Two great steps that work better together.
\newblock {\em arXiv preprint arXiv:2407.10930}, 2024.

\bibitem{srivastava2022beyond}
Aarohi Srivastava, Abhinav Rastogi, Abhishek Rao, Abu Awal~Md Shoeb, Abubakar Abid, Adam Fisch, Adam~R Brown, Adam Santoro, Aditya Gupta, Adri{\`a} Garriga-Alonso, et~al.
\newblock Beyond the imitation game: Quantifying and extrapolating the capabilities of language models.
\newblock {\em arXiv preprint arXiv:2206.04615}, 2022.

\bibitem{suzgun2022challenging}
Mirac Suzgun, Nathan Scales, Nathanael Sch{\"a}rli, Sebastian Gehrmann, Yi Tay, Hyung~Won Chung, Aakanksha Chowdhery, Quoc~V Le, Ed~H Chi, Denny Zhou, et~al.
\newblock Challenging big-bench tasks and whether chain-of-thought can solve them.
\newblock {\em arXiv preprint arXiv:2210.09261}, 2022.

\bibitem{wang2024correctly}
Wenyi Wang, Hisham~A Alyahya, Dylan~R Ashley, Oleg Serikov, Dmitrii Khizbullin, Francesco Faccio, and J{\"u}rgen Schmidhuber.
\newblock How to correctly do semantic backpropagation on language-based agentic systems.
\newblock {\em arXiv preprint arXiv:2412.03624}, 2024.

\bibitem{wei2022chain}
Jason Wei, Xuezhi Wang, Dale Schuurmans, Maarten Bosma, Fei Xia, Ed Chi, Quoc~V Le, Denny Zhou, et~al.
\newblock Chain-of-thought prompting elicits reasoning in large language models.
\newblock {\em Advances in neural information processing systems}, 35:24824--24837, 2022.

\bibitem{yang2023large}
Chengrun Yang, Xuezhi Wang, Yifeng Lu, Hanxiao Liu, Quoc~V Le, Denny Zhou, and Xinyun Chen.
\newblock Large language models as optimizers.
\newblock {\em arXiv preprint arXiv:2309.03409}, 2023.

\bibitem{yang2018hotpotqa}
Zhilin Yang, Peng Qi, Saizheng Zhang, Yoshua Bengio, William~W Cohen, Ruslan Salakhutdinov, and Christopher~D Manning.
\newblock Hotpotqa: A dataset for diverse, explainable multi-hop question answering.
\newblock {\em arXiv preprint arXiv:1809.09600}, 2018.

\bibitem{yao2022react}
Shunyu Yao, Jeffrey Zhao, Dian Yu, Nan Du, Izhak Shafran, Karthik Narasimhan, and Yuan Cao.
\newblock React: Synergizing reasoning and acting in language models.
\newblock {\em arXiv preprint arXiv:2210.03629}, 2022.

\bibitem{yuksekgonul2024textgrad}
Mert Yuksekgonul, Federico Bianchi, Joseph Boen, Sheng Liu, Zhi Huang, Carlos Guestrin, and James Zou.
\newblock Textgrad: Automatic" differentiation" via text.
\newblock {\em arXiv preprint arXiv:2406.07496}, 2024.

\bibitem{zhang2022tempera}
Tianjun Zhang, Xuezhi Wang, Denny Zhou, Dale Schuurmans, and Joseph~E Gonzalez.
\newblock Tempera: Test-time prompting via reinforcement learning.
\newblock {\em arXiv preprint arXiv:2211.11890}, 2022.

\bibitem{zheng2023judging}
Lianmin Zheng, Wei-Lin Chiang, Ying Sheng, Siyuan Zhuang, Zhanghao Wu, Yonghao Zhuang, Zi Lin, Zhuohan Li, Dacheng Li, Eric Xing, et~al.
\newblock Judging llm-as-a-judge with mt-bench and chatbot arena.
\newblock {\em Advances in Neural Information Processing Systems}, 36:46595--46623, 2023.

\bibitem{zhou2022large}
Yongchao Zhou, Andrei~Ioan Muresanu, Ziwen Han, Keiran Paster, Silviu Pitis, Harris Chan, and Jimmy Ba.
\newblock Large language models are human-level prompt engineers.
\newblock {\em arXiv preprint arXiv:2211.01910}, 2022.

\end{thebibliography}
}

\appendix

\section{Meta Prompts}

\minisection{Meta Prompts for Backward Engines(GradComponent)} \label{appendix: meta_prompts_backward_engine}

This includes Generator, Retriever and any developer subclassed GradComponent.

Most of these prompts are located in directory:\\ \texttt{adalflow/adalflow/optim/text\_grad/backend\_engine\_prompt.py}.

\begin{lstlisting}[language=Python, caption={Backward engine template.},label={lst:meta_prompt_backward_template}]

FEEDBACK_ENGINE_TEMPLATE = r"""<START_OF_SYSTEM_PROMPT>
You MUST determining the root cause of a system error.
You start with an evaluation function that measures performance, and you receive the system input.
The system can be a a compound system, potentially consisting of multiple components.
You work on one component.
You will receive feedback from your direct successor component, and your goal is to investigate your component’s inputs and outputs to identify whether any of your input variables are causing the error.

Your target input variable is enclosed in <TARGET_VARIABLE> (representing one of the input variables that may or may not be causing the error).
Alternatively, it may be enclosed in <VARIABLES> tags (in which case you must pass feedback to all variables, indicating which ones cause the errors and which do not).

1. From <CONVERSATION></CONVERSATION> section, you can find how the variable is obtained and used.
2. As there might be multiple precedessors, and multi-components, it is possible that the feedback/error is not directly related to the variable itself.
3. When you reason, really think about the variable's role in the component(infer from the CONVERSATION section) and the VARIABLE section before you provide feedback.
4. Be specific, concise, critical, and direct.
5. Maximum 3 sentences.

[Cycle]: If the same DataID has multiple gradients, it means this component/variable is called multiple times in the compound system(with a cycle) in the same order as it appears in the gradient list.
   Ensure the feedback is aware of all sets of inputs and outputs.

{% if output_format_str %}
{{output_format_str}}
{% endif %}

<END_OF_SYSTEM_PROMPT>
<START_OF_USER>
<CONVERSATION>
{{conversation_sec}}
</CONVERSATION>
<OBJECTIVE_INSTRUCTION>
{{objective_instruction_sec}}
</OBJECTIVE_INSTRUCTION>
<END_OF_USER>
"""
\end{lstlisting}

\begin{lstlisting}[language=Python, caption={Intermediate objective template(General GradComponent).}, label={lst:intermediate_objective_grad_component_template}]
OBJECTIVE_INSTRUCTION_CHAIN = r"""This conversation is part of a larger system. The <INPUTS/SCORE> was later used as "{{response_name}}: {{response_desc}}".
<OBJECTIVE_FUNCTION>
Your only goal is to clearly states how it obtained the "Eval output/score": {{response_gradient}}.
Especially when the score is low.
Be CONCISE.
If you have enough context, add more specific feedback on how it failed.
e.g. "The retrieved context is not enough to answer the question so the problem relies on the retrieval part."
</OBJECTIVE_FUNCTION>"""
\end{lstlisting}

\begin{lstlisting}[language=Python, caption={Intermediate objective template(Generator).}, label={lst:intermediate_objective_generator_template}]
OBJECTIVE_INSTRUCTION_CHAIN = r"""
This component is part of a larger system. The <LM_OUTPUT> was later used as {{response_desc}}.
<OBJECTIVE_FUNCTION>
Your goal is to give feedback to the variable to guide the LLM_OUTPUT according to feedback: {{response_gradient}}
{% if instruction_to_backward_engine %}
Note: {{instruction_to_backward_engine}}
{% endif %}
</OBJECTIVE_FUNCTION>"""
\end{lstlisting}

\begin{lstlisting}[language=Python, caption={LLM conversation template.}, label={lst:llm_conversation_template}]
CONVERSATION_START_INSTRUCTION_CHAIN = r"""
{{variable_and_peers_info}}

{# system trainable variables #}
{% if predecessors %}
<START_OF_PREDECESSORS>
The target variable is used together with these predecessors variables besides of the peers:
{% for system_variable in predecessors %}
{{loop.index}}.
Name: {{system_variable.name}}
Type: {{system_variable.param_type}}
Description: {{system_variable.role_desc}}
WILL_BE_OPTIMIZED: {{system_variable.requires_opt}}
Value: {{system_variable.prompt_data}}
{% endfor %}
<END_OF_PREDECESSORS>
{% endif %}

Here is the inputs and output with this component(LM):
{{conversation_str}}
"""

LLM_CONVERSATION_TEMPLATE = r"""
LM_INPUT: {{input_value}}
LM_OUTPUT: {{llm_output}}
{% if gt %}
GROUND_TRUTH: {{gt}}
{% endif %}
"""
\end{lstlisting}

\begin{lstlisting}[language=Python, caption={LLM's Variables and Peer template.}, label={lst:llm_variables_peers_template}]
VARIABLE_AND_PEERS_INFO = r"""
<START_OF_VARIABLE_DESC>
<NAME> {{variable.name}} </NAME>
<TYPE> {{variable.param_type}} </TYPE>
<ROLE> {{variable.role_desc}} </ROLE>
<VARIABLE>{{ variable.prompt_data}}</VARIABLE>
<END_OF_VARIABLE_DESC>
{% if peers %}
<VARIBLE_PEERS>
{% for peer in peers %}
{{loop.index}}.
PEER_NAME: {{peer.name}},
PEER_TYPE: {{peer.param_type}},
PEER_ROLE: {{peer.role_desc}}
WILL_BE_OPTIMIZED: {{peer.requires_opt}}
{% if peer.prompt_data %}
PEER_VARIABLE: {{peer.prompt_data}}
{% else %}
PEER_VARIABLE: EMPTY
{% endif %}
{% endfor %}
</VARIBLE_PEERS>
{% endif %}
"""
\end{lstlisting}

\begin{lstlisting}[language=Python, caption={Intermediate conversation template(GradComponent).}, label={lst:intermediate_conversation_grad_component_template}]
OBJECTIVE_INSTRUCTION_CHAIN = r"""This conversation is part of a larger system. The <INPUTS/SCORE> was later used as "{{response_name}}: {{response_desc}}".
<OBJECTIVE_FUNCTION>
Your only goal is to clearly state how it obtained the "Eval output/score": {{response_gradient}}.
Especially when the score is low.
Be CONCISE.
If you have enough context, add more specific feedback on how it failed.
e.g. "The retrieved context is not enough to answer the question so the problem relies on the retrieval part."
</OBJECTIVE_FUNCTION>"""

GRAD_COMPONENT_CONVERSATION_TEMPLATE_STRING = r"""
COMPONENT_DESC: {{component_desc}}

INPUTS:
{% for key, (value, eval_type) in inputs.items() %}
{{loop.index}}.
KEY: {{ key }}.
ROLE: {{ value.role_desc }},
DATA: {{ value.prompt_data }},
{% endfor %}

OUTPUT: {{response_value}}
{% if metadata %}
Note: {{metadata}}
{% endif %}"""
\end{lstlisting}

\minisection{Meta-Prompts for LossComponent}

Code for LossComponent is at: \\
\texttt{adalflow/adalflow/optim/text\_grad/text\_loss\_with\_eval\_fn.py}.

\begin{lstlisting}[language=Python, caption={Loss Objective template.}, label={lst:loss_objective_template}]

LOSS_CONVERSATION_START_INSTRUCTION_STRING_FN = r"""
TARGET VARIABLE:
<NAME> {{variable.name}} </NAME>
<ROLE> {{variable.role_desc}} </ROLE>
<VARIABLE> {{variable.prompt_data}} </VARIABLE>
{{conversation_str}}
"""

# for conversation_str
LOSS_CONVERSATION_TEMPLATE_STRING = r"""
The variable is passed to the eval function and compared with a target/ground truth value to get
its score regarding to a SYSTEM_QUESTION: {{system_question}}.

EVAL_FUNC: {{eval_fn_desc}}

INPUTS to EVAL_FUNC:
{% for key, (value, eval_type) in inputs.items() %}
({{ key }}) (role: {{ value.role_desc }}),
data: {{ value.prompt_data }},
input_to_eval_fn: {{ value.eval_input }},
data_type: {{ eval_type }}
{% endfor %}

OUTPUTS/SCORE: {{response_value}}
{% if metadata %}
Note: {{metadata}}
{% endif %}"""

# for objective_str
OBJECTIVE_INSTRUCTION_BASE = r"""<OBJECTIVE_FUNCTION>
Your task is to provide the response with specific feedback based on the expected correct response (y_gt/ground_truth) and the score in the "<OUTPUTS/SCORE>".
Especially when the score is low.
Be CONCISE.

Be specific on why it has a low score.
Specify the difference between the expected correct response and the response.
</OBJECTIVE_FUNCTION>"""
"""
\end{lstlisting}

\minisection{Meta prompts for the optimizer}

The optimizer is located at:\\
\texttt{adalflow/adalflow/optim/text\_grad/tgd\_optimizer.py}

\begin{lstlisting}[language=Python, caption={Gradient Driven LLM Optimizer(GDLO) template.}, label={lst:gdlo_template}]
TEXT_GRAD_DESC_TEMPLATE = r"""<START_OF_SYSTEM_PROMPT>
{{optimizer_system_prompt}}
<END_OF_SYSTEM_PROMPT>
<START_OF_USER_MESSAGE>
You are {{steps}} steps since your last improvement.
Update the value more rapidly when steps are larger than 3.
{# Variable and peers info #}
<START_OF_VARIABLE_AND_PEERS_INFO>
{{variable_and_peers_info}}
<END_OF_VARIABLE_AND_PEERS_INFO>
{# system trainable variables #}
{% if system_variables %}
<START_OF_SYSTEM_VARIABLES>
The target variable is used together with these system variables besides of its peers:
{% for system_variable in system_variables %}
{{loop.index}}.
Name: {{system_variable.name}}
Type: {{system_variable.param_type}}
Description: {{system_variable.role_desc}}
WILL_BE_OPTIMIZED: {{system_variable.requires_opt}}
Value: {{system_variable.prompt_data}}
{% endfor %}
Strategically plan the role of each system variable to collaborate with each other for final correct answer.
<END_OF_SYSTEM_VARIABLES>
{% endif %}
{# OPRO past history #}
{% if past_history %}
<START_OF_HISTORY_PERFORMANCE>
Here are the best past iterations.
{% for history in past_history %}
{{loop.index}}. {{history}}
{% endfor %}
IMPORTANT: Your goal is to generate new variable that score higher than all past iterations.
<END_OF_HISTORY_PERFORMANCE>
{% endif %}
{# Multi-proposal history #}
{% if failed_proposals %}
<START_OF_CURRENT_ITERATION>
same batch, same feedback: Here are the values you have tried that have not improved the score.(scored <= {{best_score}}):
{% for failed_proposal in failed_proposals %}
{{loop.index}}. {{failed_proposal}}
{% endfor %}
You MUST approach differently from the above methods.
<END_OF_CURRENT_ITERATION>
{% endif %}
{# Feedback #}
{% if variable_grad %}
<START_OF_CONTEXT_FEEDBACK>
Here are the context and feedback for the variable:
{{variable_grad}}
<END_OF_CONTEXT_FEEDBACK>
{% endif %}
{# Constraints #}
{% if constraint_text %}
You must follow the following constraints:
<CONSTRAINTS>{{constraint_text}}</CONSTRAINTS>
{% endif %}
{# In-context examples #}
{% if in_context_examples %}
You must base on the following examples when modifying the {{variable_desc}}:
<EXAMPLES>{{in_context_examples}}</EXAMPLES>
{% endif %}
<END_OF_USER_MESSAGE>
"""
\end{lstlisting}

In AdalFlow, the gradient context is tracked as Data Structure \textit{GradientContext} in the backward pass along with the gradients.
In the forward pass, we use similr code to track all context
\begin{lstlisting} [language=Python, basicstyle=\ttfamily\small, breaklines=true]
    pred.gradients_context[var_gradient] = GradientContext(
            context=conversation_str,
            response_desc=response.role_desc,
            parameter_desc=pred.role_desc,
        )
\end{lstlisting}
The goal is to set the context for predecessor node.

\section{Experimental implementation} \label{appendix:experimental_implementation}
Here are the hyperparameters used for training:
\begin{lstlisting}
gpt_3_model = {
    "model_client": OpenAIClient(input_type="text"),
    "model_kwargs": {
        "model": "gpt-3.5-turbo-0125",
        "max_tokens": 2000,
        "temperature": 0.0,
        "top_p": 0.99,
        "frequency_penalty": 0,
        "presence_penalty": 0,
        "stop": None,
    },
}
gpt_4o_model = {
    "model_client": OpenAIClient(),
    "model_kwargs": {
        "model": "gpt-4o",
        "temperature": 1,
        "top_p": 0.99,
    },
}
\end{lstlisting}
\section{Application patterns}

\paragraph{ObjectCount}
The runtime computation graph:
\begin{figure} [H]
    \centering
    \includegraphics[width=0.4\textwidth]{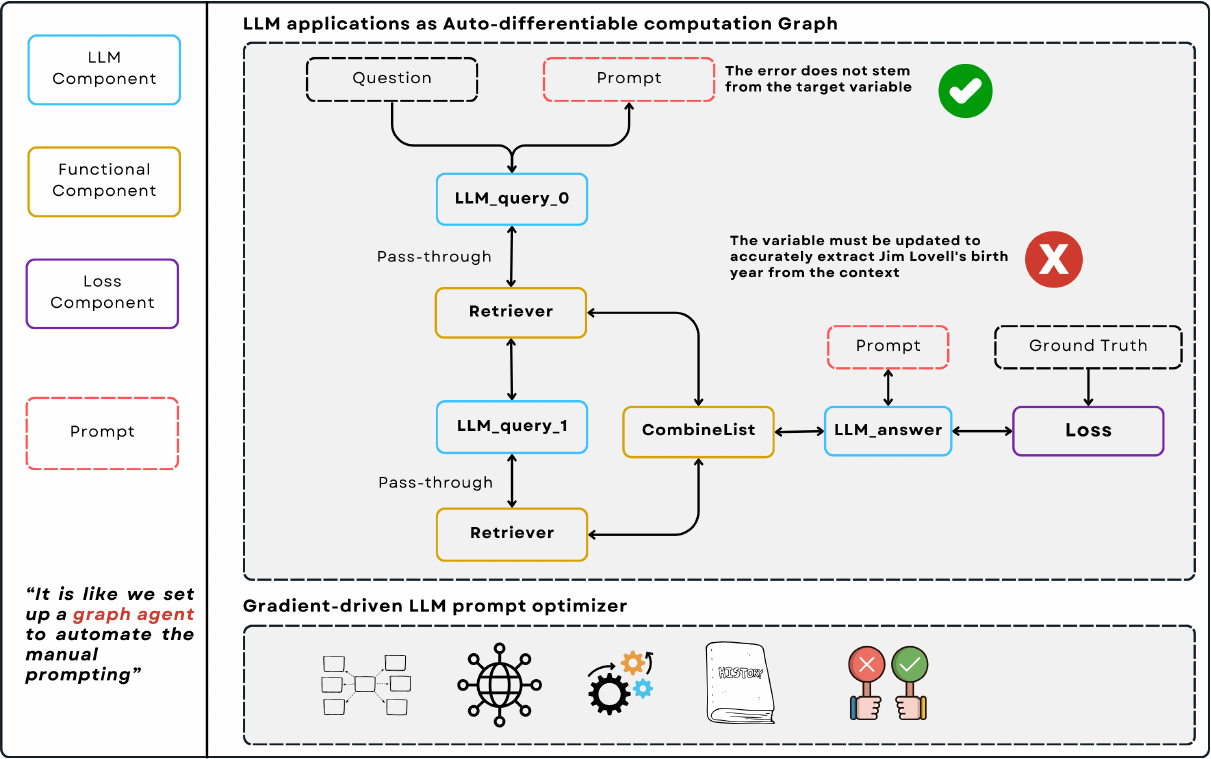}
    
    \caption{Object count computation graph}
    \label{fig:object_count}
\end{figure}
\paragraph{Trec-10}
The runtime computation graph:
\begin{figure} [H]
    \centering
    \includegraphics[width=1\textwidth]{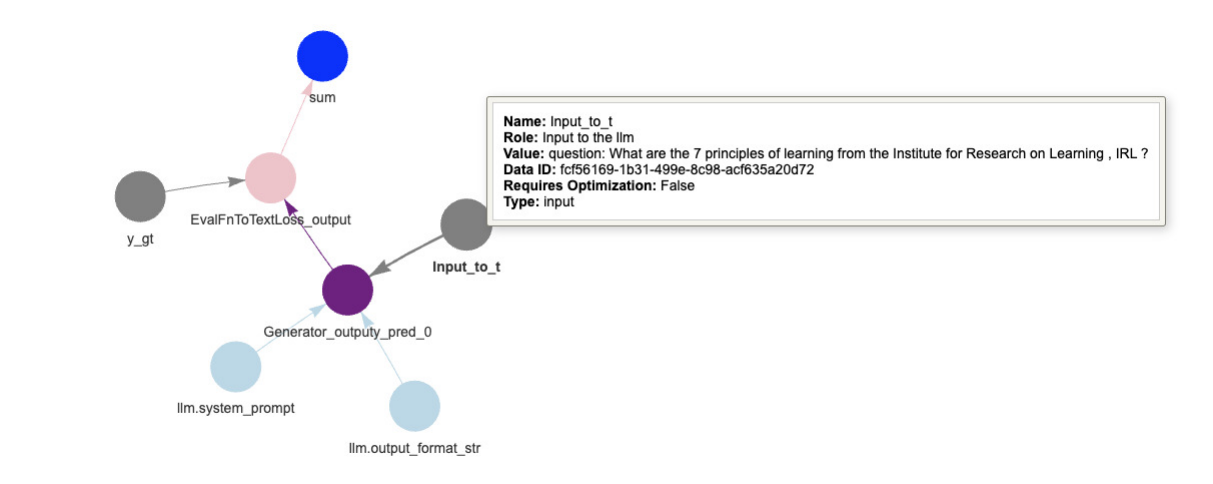}
    \caption{Trec-6 computation graph}
    \label{fig:trec_6}
\end{figure}

\paragraph{Vanilla RAG}
The runtime computation graph:
\begin{figure} [H]
    \centering
    \includegraphics[width=0.8\textwidth]{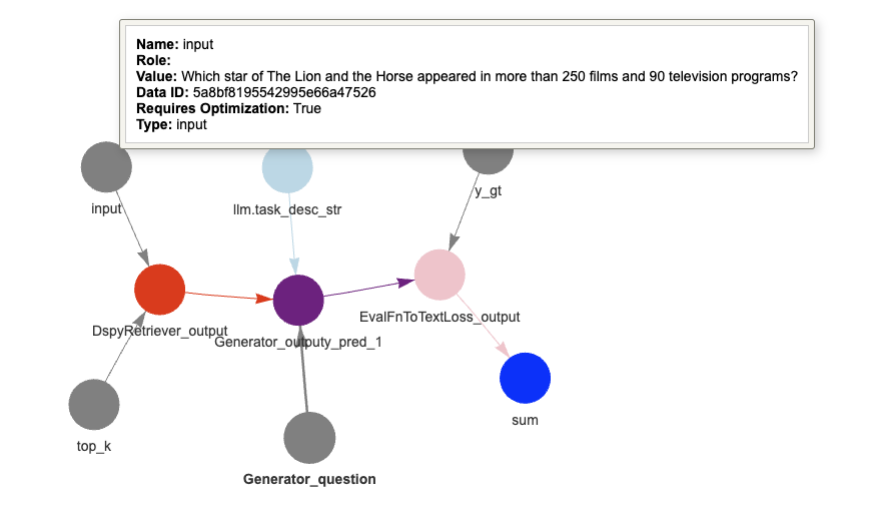}
    \caption{Vanilla RAG computation graph.}
    \label{fig:vanilla_rag}
\end{figure}

\paragraph{Multi-hop RAG}
The runtime computation graph:
\begin{figure} [H]
    \centering
    \includegraphics[width=1\textwidth]{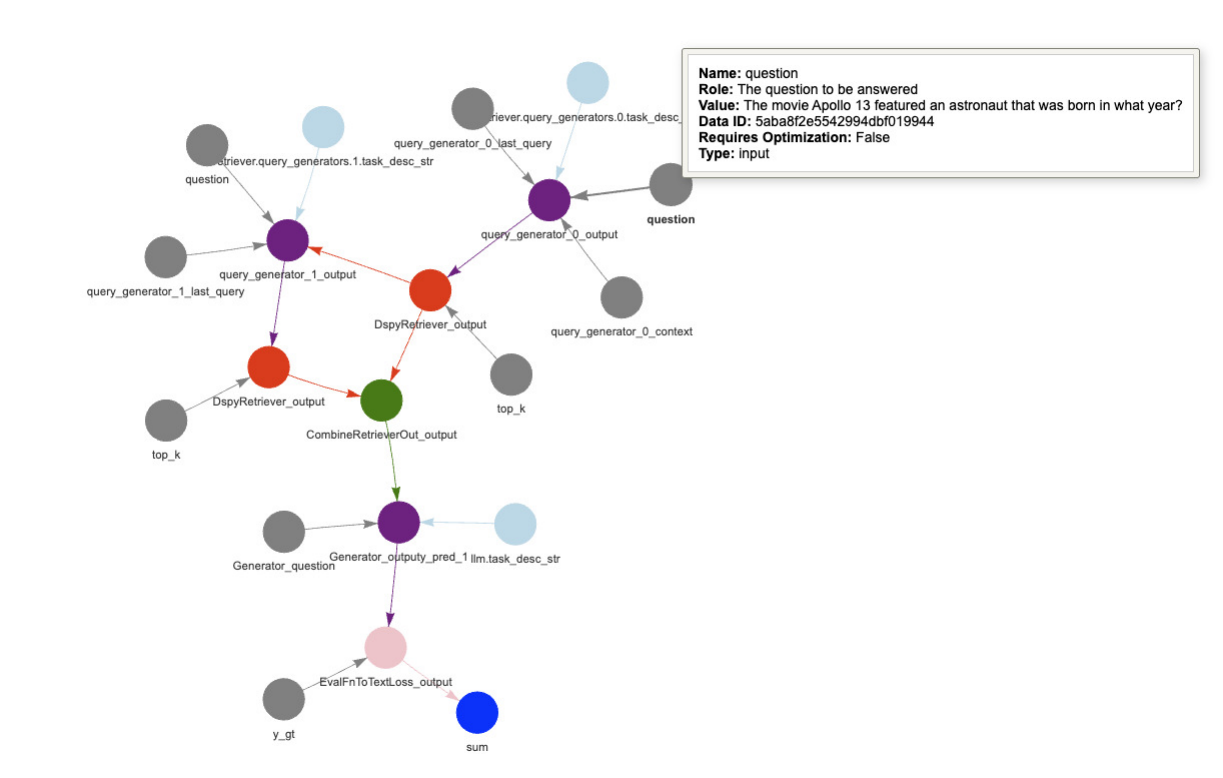}
    \caption{Multi-hop RAG computation graph. We see \textit{query\_generator\_0\_output} and \textit{query\_generator\_1\_output} are outputs from two different genertor components.}
    \label{fig:multihop_rag}
\end{figure}

\paragraph{Multi-hop RAG(Cycle)}
The runtime computation graph:
\begin{figure} [H]
    \centering
    \includegraphics[width=1\textwidth]{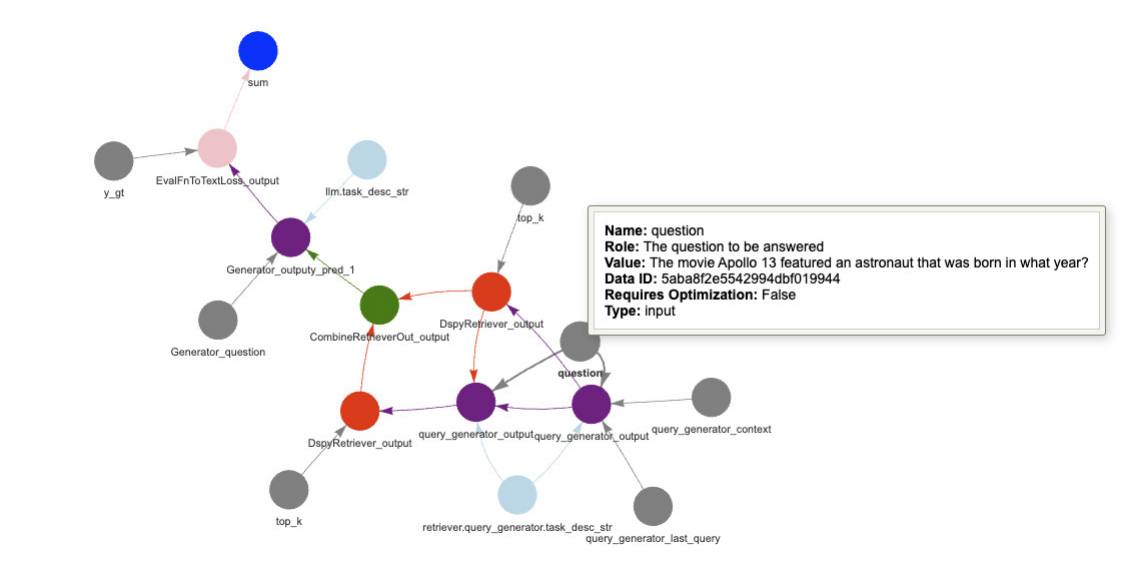}
    \caption{Multi-hop RAG(Cycle) computation graph. We see \textit{query\_generator\_output} as the output of one generator component but being called twice in two sequential steps which forms a cycle in the component graph.}
    \label{fig:multihop_rag_cycle}
\end{figure}

\paragraph{Agentic RAG}

\paragraph{Code and files} As shown in Tab. ~\ref{tab:methods_code_resources}.

To run the experiments, use script: \texttt{use\_cases/bmp\_train.py}.
\begin{table*}[ht!]
    \centering
    \tiny
    \begin{tabularx}{\textwidth}{l X X}
        \toprule
        \textbf{Method / App}
         & \textbf{Code (GitHub)}
         & \textbf{Graph Dir}                                                           \\
        \midrule

        \textbf{ObjectCount}
         & use\_cases/question\_answering/bbh/object\_count/train\_new.py
         & fig/object\_count\_runtime\_graph/                                           \\
        \midrule
        \textbf{Trec-10}
         & use\_cases/classification/train.py
         & fig/trec\_6\_runtime\_graph/
        \\
        \midrule

        \textbf{Vanilla RAG}
         & benchmarks/hotpot\_qa/adal\_exp/train\_vanilla\_rag.py

         & fig/vanilla\_rag\_runtime\_graph/                                            \\
        \midrule


        \textbf{Multi-hop RAG}
         & benchmarks/hotpot\_qa/adal\_exp/train\_multi\_hop\_rag.py
         & fig/multi\_hop\_rag\_runtime\_graph/                      \\

                \textbf{Multi-hop RAG(Cycle)}
         & benchmarks/hotpot\_qa/adal\_exp/train\_multi\_hop\_rag\_cycle.py
         & fig/multi\_hop\_rag\_cycle\_runtime\_graph/                     \\

      \textbf{Agentic RAG}
         & benchmarks/hotpot\_qa/adal\_exp/train\_agent\_rag.py
         &                     \\

        \bottomrule
    \end{tabularx}
    \caption{Methods/Applications with corresponding code and resource links.}
    \label{tab:methods_code_resources}
\end{table*}

\subsection{Baselines}

\paragraph{DsPy MIPROv2}

Here is the code. We use the same split of data and the same prompt, except the output format as DsPy controls it internally.

\begin{lstlisting}[
    basicstyle=\ttfamily\small,
    breaklines=true,
    language=Python,
    keywordstyle=\color{blue}\bfseries,
    stringstyle=\color{orange},
    commentstyle=\color{green!50!black},
    frame=single,
    showstringspaces=false,
    tabsize=4,
    numbers=left,
    numberstyle=\tiny\color{gray}]
def train_MIPROv2(trainset, valset, save_path, filename):

import os
from dspy.teleprompt import MIPROv2

if not os.path.exists(save_path):
    os.makedirs(save_path)

tp = MIPROv2(
    metric=validate_exact_match,
    prompt_model=gpt_4,
    task_model=turbo,
    num_candidates=30,
    init_temperature=1.0,
)
compiled_task = tp.compile(
    ObjectCount(),
    trainset=trainset,
    valset=valset,
    max_bootstrapped_demos=5,
    max_labeled_demos=2,
    num_batches=12,  # MINIBATCH_SIZE = 25,
    seed=2025,
    requires_permission_to_run=False,
)
compiled_task.save(os.path.join(save_path, filename))
return compiled_task
\end{lstlisting}

\end{document}